\documentclass[sigconf, nonacm]{acmart}
\renewcommand\footnotetextcopyrightpermission[1]{}
\AtBeginDocument{%
  \providecommand\BibTeX{{%
    \normalfont B\kern-0.5em{\scshape i\kern-0.25em b}\kern-0.8em\TeX}}}

\usepackage[T1]{fontenc}
\usepackage{aecompl}
\usepackage{amsmath}

\usepackage{amssymb}
\usepackage{pifont}
\usepackage{bm}
\usepackage{multirow}
\usepackage{subfigure}
\usepackage{algorithm}
\usepackage{algpseudocode}
\usepackage{siunitx}
\usepackage{color}
\usepackage{colortbl}

\begin{document}
\title{Hierarchical Knowledge Guided Fault Intensity Diagnosis of Complex Industrial Systems}

% \affiliation{%
\author{Yu Sha$^{1,3,4}$, Shuiping Gou$^{1,\text{\dag}}$, Bo Liu$^{1}$, Johannes Faber$^{3}$, Ningtao Liu$^{1}$, \\Stefan Schramm$^{3}$, Horst Stoecker$^{3,5,6}$, Thomas Steckenreiter$^{7}$, Domagoj Vnucec$^{7}$, \\Nadine Wetzstein$^{7}$, Andreas Widl$^{7}$ and Kai Zhou$^{2,3, \text{\dag}}$\\
{$^1$School of Artificial Intelligence, Xidian University, Xian, China \\
$^2$School of Science and Engineering, The Chinese University of Hong Kong, Shenzhen, China \\
$^3$Frankfurt Institute for Advanced Studies, Frankfurt am Main, Germany \\ 
$^4$Xidian-FIAS international Joint Research Center, Frankfurt am Main, Germany \\
$^5$Institut f{\"u}r Theoretische Physik, Goethe Universit{\"a}t Frankfurt, Frankfurt am Main, Germany \\
$^6$GSI Helmholtzzentrum f{\"u}r Schwerionenforschung GmbH, Darmstadt, Germany \\
$^7$SAMSON AG, Frankfurt am Main, Germany}}

%% You do not have to enter your paper ID

\renewcommand{\shortauthors}{Yu Sha et al.}

\begin{abstract}
Fault intensity diagnosis (FID) plays a pivotal role in monitoring and maintaining mechanical devices within complex industrial systems. As current FID methods are based on chain of thought without considering dependencies among target classes. To capture and explore dependencies, we propose a \underline{h}ierarchical \underline{k}nowledge \underline{g}uided fault intensity diagnosis framework (HKG) inspired by the tree of thought, which is amenable to any representation learning methods. The HKG uses graph convolutional networks to map the hierarchical topological graph of class representations into a set of interdependent global hierarchical classifiers, where each node is denoted by word embeddings of a class. These global hierarchical classifiers are applied to learned deep features extracted by representation learning, allowing the entire model to be end-to-end learnable. In addition, we develop a re-weighted hierarchical knowledge correlation matrix (Re-HKCM) scheme by embedding inter-class hierarchical knowledge into a data-driven statistical correlation matrix (SCM) which effectively guides the information sharing of nodes in graphical convolutional neural networks and avoids over-smoothing issues. The Re-HKCM is derived from the SCM through a series of mathematical transformations. Extensive experiments are performed on four real-world datasets from different industrial domains (three cavitation datasets from SAMSON AG and one existing publicly) for FID, all showing superior results and outperform recent state-of-the-art FID methods.
\end{abstract} 

\ccsdesc[500]{Computing methodologies~Machine learning}
\ccsdesc[300]{Computing methodologies~Machine learning approaches}
\ccsdesc{Computing methodologies~Representation learning}
\ccsdesc{Computing methodologies~Graph convolutional network}

% \vspace{-5pt}
\keywords{Cavitation Intensity Diagnosis; Acoustic Signals; Hierarchical Knowledge; Hierarchical classification; Representation Learning and Graph Convolutional Network}



\maketitle
\renewcommand{\thefootnote}{}
\footnotetext{$^\text{\dag}$Corresponding authors}

\section{Introduction}
\label{sec:intro}
Deep learning has achieved impressive performance in numerous signal processing applications, such as fault detection, fault location and fault intensity diagnosis (FID) \cite{hundman2018detecting}. In this paper, we focus on FID for cavitation or other faults in complex industrial mechanical systems, which has been an active research topic in the industry during recent years. The essence of FID is to pinpoint the finer fault within the acoustic or vibration signals emitted by the target machine, which is regarded as a fundamental technology in the fourth industrial revolution \cite{munirathinam2020industry}. Monitoring the health of a machine can make a huge contribution to the overall maintenance and efficiency of industrial processes by 'listening' to signals emitted from the machine \cite{farrar2012structural}.

Formally, FID data comprises multiple measurement signals, each describing the whole physical process from the beginning to the end of a particular fault state arising in a complex entity. Therefore, there is a natural interdependency among each measurement event. This interdependency is also referred to as hierarchical knowledge between measurement events, which can be systematically organized into a hierarchical knowledge tree \cite{li2022deep}. Disregarding this interdependency may lead to degradation of fault intensity diagnosis performance. In order for the mechanical equipment health monitoring system to work better in complex industrial systems, the algorithm should take full account of interdependencies between measurement events, i.e., hierarchical knowledge.

Recent FID methods can be broadly categorized into two classes based on convolutional neural networks (CNN) and Transformer. CNN-based methods \cite{mohammad2023one,zhao2020deep,pan2017liftingnet} use filters (i.e., convolutional structures) to capture spatio-temporal features from vibration or acoustic signals. Then, multiple-layer convolutional structures are implemented to model complex patterns of signals to achieve accurate fault diagnosis. For transformer-based approaches \cite{yu2023adaptive,cui2024self}, they explore long-term dependencies across signal sequences through a self-attention mechanism, allowing a better understanding of the intrinsic correlations between the signals. This approach is good at dealing with non-stationary signals and complex relationships.

\begin{figure}
    \centering
    \includegraphics[width=0.3\textwidth,height=33mm]{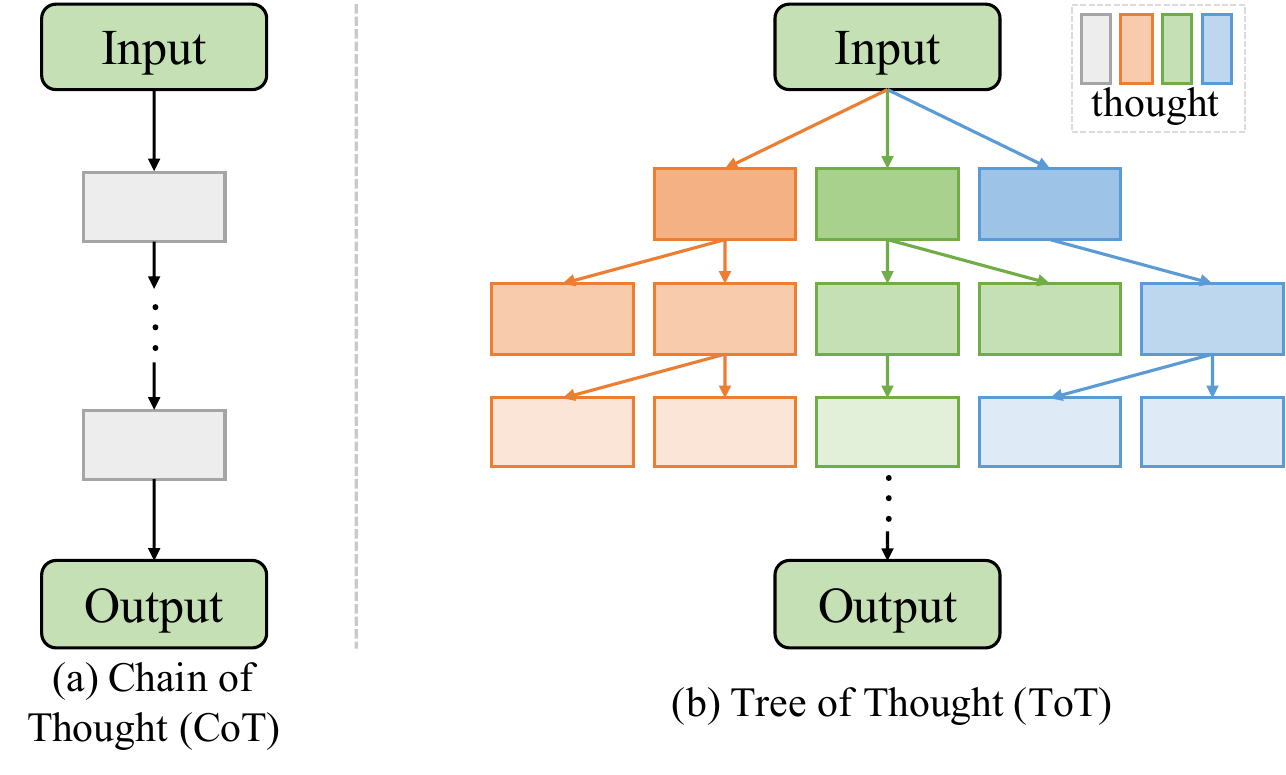}
    \caption{Schematic diagrams for visualising different thoughts. Each rectangular box represents a thought, which is an intermediate step towards the problem. Different coloured rectangular boxes of the same colour scheme indicate their associations.}
    \label{fig: DifferentThoughts}
\end{figure}

However, CNN- or Transformers-based methods apart from their own limitations, common drawback is that they are based on the traditional input-output chain of thought (CoT) \cite{wei2022chain}, as shown in Figure \ref{fig: DifferentThoughts}a. CoT treats the problem as a direct mapping between inputs and outputs and neglects implicit relationships (i.e., hierarchical relationships) between target classes, which affects the performance and generalisability of the model. To overcome this limitation, we suggest a tree of thought (ToT) paradigm \cite{yao2023tree}, i.e., embedding hierarchical knowledge between target classes into the CoT, see Figure \ref{fig: DifferentThoughts}b. Each node of the ToT represents a specific fault state and edges denote the hierarchical relationships among nodes. The ToT paradigm is consistent with the inherent hierarchical nature of fault data, which can capture hierarchical dependencies across fault classes and provide a richer representation of the relationships between various fault states to guide the model towards a more nuanced understanding of fault classes.

Based on the above considerations, for industrial signals fault intensity diagnosis, we propose to explicitly embed hierarchical knowledge among target classes into the current deep learning methods (i.e., CNNs and Transformers) to better enhance the ability of the model to understand the complex inter-dependencies and intrinsic structure across fault data. To convert CoT into ToT, instead of previous structured FID models focusing on sophisticated network design, we directly formulate FID as a feature multi-label classification task. This allows hierarchical knowledge to be easily embedded into any current representation learning (ML) methods and better guides learned deep features from ML. To make feature classification consistent with the hierarchical structure of classes, we use graph convolutional networks to map the hierarchical topology of class/label representations along with the learned deep features from ML into interdependent target classifiers, i.e., hierarchical classifiers. Furthermore, the hierarchical classifier with hierarchical constraints is not a set of independently trained parameter vectors, but it implements end-to-end learning with learned deep features. The hierarchical classifier is also a global classifier since its mapping parameters are common to all target classes. In addition, we develop a re-weighted hierarchical knowledge correlation matrix (Re-HKCM) paradigm by embedding the hierarchical knowledge across classes into a data-driven statistical correlation matrix (SCM), which can explicitly model inter-class dependencies and alleviate the over-smoothing problem. The Re-HKCM is derived from the SCM through a series of mathematical transformations. The above forms our proposed knowledge- and data-driven approach, namely \underline{H}ierarchical \underline{K}nowledge \underline{G}uided (HKG) fault intensity diagnosis in complex industrial systems.

The contributions of this paper are summarized as follows:
\begin{itemize}
    \item A novel end-to-end \underline{h}ierarchical \underline{k}nowledge-\underline{g}uided fault intensity diagnosis framework (HKG) is proposed, which can be employed to any representation learning approaches.
    \item We develop a global hierarchical classifier using graph convolutional neural networks which maps the hierarchical topological graph of the label representation to the inter-dependent target classifiers.
    \item We design a re-weighted hierarchical knowledge correlation matrix scheme by embedding inter-class hierarchical knowledge into a data-driven statistical correlation matrix, which efficiently constrains learned deep features and mitigates the phenomenon of over-smoothing. 
    \item Our HKG outperforms state-of-the-art methods in experiments conducted on four real-world datasets from different industrial domains (three cavitation datasets provided by SAMSON AG and one public dataset). Moreover, ablation studies further demonstrate the effectiveness of our proposed HKG for industrial fault intensity diagnosis. 
\end{itemize}

\section{PRELIMINARIES}
\label{sec: preliminaries}

\subsection{Cavitation Event Intensity Recognition}
\label{sec: Cavitation Intensity Recognition}
Cavitation is defined as the phenomenon of the formation, growth and collapse of local bubbles or vapor cavities in a liquid \cite{plesset1977bubble}. For piping systems, the acoustic signals of different flow conditions (different levels of cavitation or non-cavitation) are recorded as continuous waveforms using acoustic sensors, as shown in Figure \ref{fig: different cavitations}. Different intensities of cavitation are defined in the Appendix. Each observed acoustic signal records the entire physical process from the beginning to the end of the event of the corresponding flow state. In our experiments, cavitation intensity recognition mainly distinguishes incipient cavitation, constant cavitation, choked flow cavitation and non-cavitation. Whether severe or subtle, any type of cavitation would indicate a potential problem or failure in system operation. Therefore, there is an urgent need for operators of industrial systems to effectively and precisely recognize different intensities of cavitation to take suitable countermeasures. 
\begin{figure}[htbp]
\centering
\subfigure[Choked Flow cavitation]{
\begin{minipage}[t]{0.44\linewidth}
\centering
\includegraphics[width=\textwidth,height=12mm]{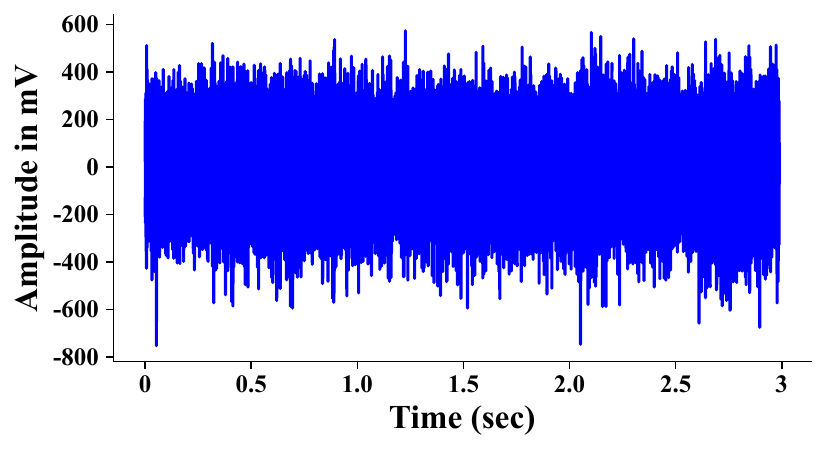}
%\caption{fig1}
\end{minipage}%
}%
\subfigure[Constant cavitation]{
\begin{minipage}[t]{0.44\linewidth}
\centering
\includegraphics[width=\textwidth,height=12mm]{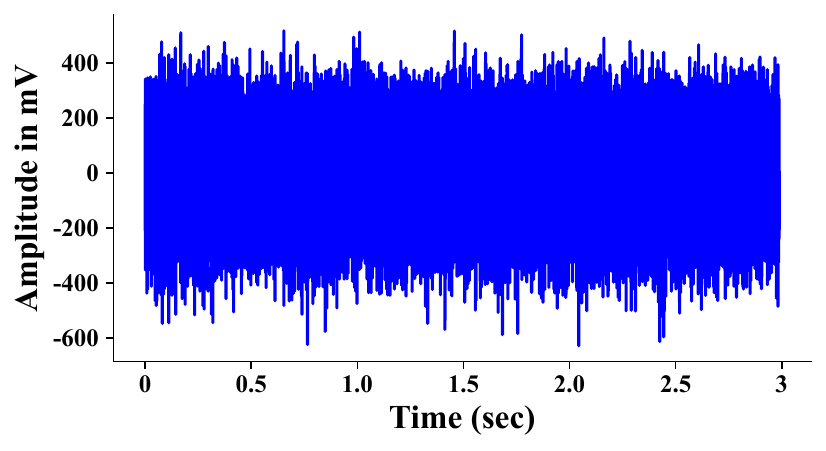}
%\caption{fig2}
\end{minipage}%
}%

\subfigure[Incipient cavitation]{
\begin{minipage}[t]{0.44\linewidth}
\centering
\includegraphics[width=\textwidth,height=12mm]{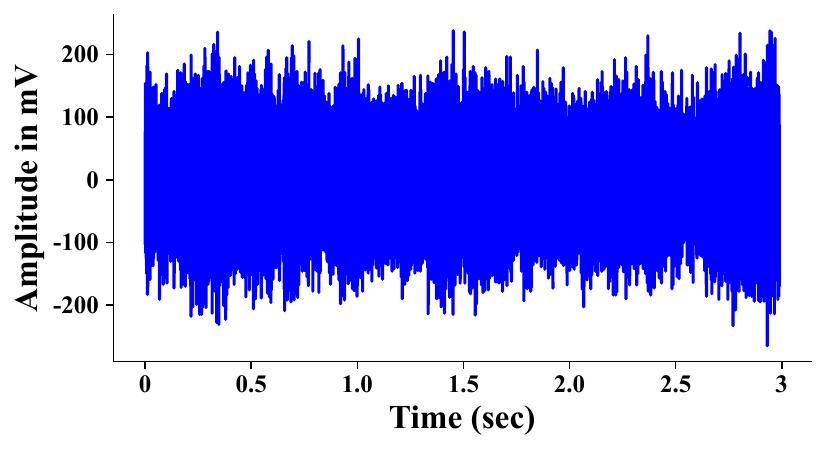}
%\caption{fig1}
\end{minipage}%
}%
\subfigure[Non cavitation]{
\begin{minipage}[t]{0.44\linewidth}
\centering
\includegraphics[width=\textwidth,height=12mm]{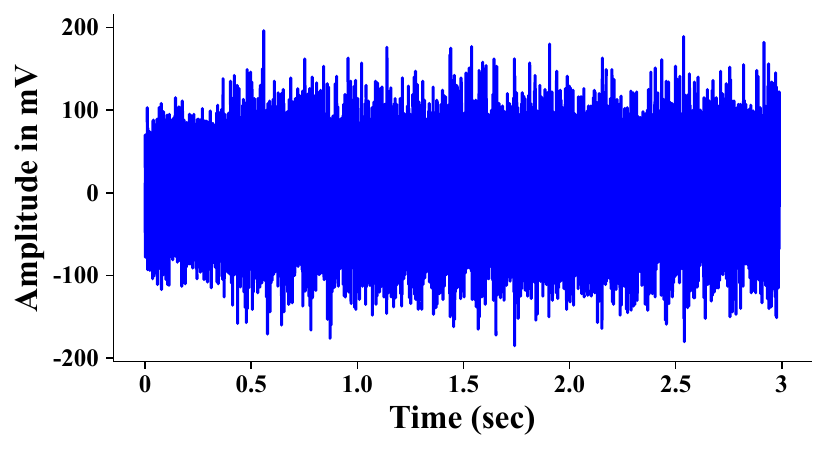}
%\caption{fig1}
\end{minipage}%
}%
\centering
\caption{Different cavitation states of acoustic signals.}
\label{fig: different cavitations}
\end{figure}

\subsection{Acoustic Signals Augmentation}
\label{sec: Acoustic Signals Augmentation}
Formally, there is $x\subseteq {\mathbb{R}}^{M\times N}$ with $M$ measurements for each acoustic signal. Considering the purposely maintained steady flow status (i.e., it’s always the same fluid status class within the individual measurement duration with \SI{3}{s} or \SI{25}{s}) in each recorded stream and fine resolution for the sensor. One can split each stream into several pieces, which still can hold enough essential information for detection. The given piece is also not too short due to the inherent randomness of the noise emission and features of each piece are independent. Therefore, we apply a sliding window (SW) with window size ${s}_{w}$ and step size ${s}_{s}$ to divide the acoustic signal sequence into a set of sub-sequences ${\mathbf{X}}_{sw}=\left\{\bm{{x}}_{i,j},i=1,2,\ldots,N;j=1,2,\ldots,k\right \}\subseteq {\mathbb{R}}^{{s}_{w}\times kN}$, where $k=\frac{\left(M-{s}_{w}\right)}{{s}_{s}}$ is the number of sub-sequences and $N$ is stream. The SW is an important part of acoustic signal pre-processing.

\subsection{Time-Frequency Transform}
\label{sec: Time-Frequency Transform}
Time-Frequency (T-F) transform provides more detailed and comprehensive information on both time and frequency dimensions \cite{abdulaal2021practical}. The most widely used T-F transform is computed by the short time Fourier transform (STFT) and can be converted back to time-domain signals by the inverse STFT (iSTFT). Given a sequence of signal $ {x[n]}_{n=0}^{N-1}$, the STFT converts the signal sequence into the T-F domain is defined by the formula:
\begin{equation}
\label{eq: STFT}
{X}_{w}[k] = \sum\limits_{n = 0}^{N - 1} {{{x}_{w}\left [n\right]}{e^{ - \frac{{2\pi j}}{N}nk}}}  := \sum\limits_{n = 0}^{N - 1} {{{x}_{w}\left [n\right]}W_N^{kn}} 
\end{equation}
where ${x}_{w}\left [n\right]=x\left [n\right]\cdot w\left [n-m\right]$ denotes the weighted signals to the window function $w\left [n-m\right]$, ${X}_{w}[k]$ is the result in the frequency domain, $j$ is the imaginary unit and ${W_N} = {e^{ - \frac{{2\pi j}}{N}}}$. The essence of the STFT is to apply a Discrete Fourier Transform (DFT) on the resulting windowed signal ${x}_{w}\left [n\right]$. The formulation of DFT of Equation \ref{eq: STFT} can be derived from the Fourier transform for continuous signals. For our method, the STFT is an essential part of the acoustic signal's pre-processing.

\subsection{Graph Convolutional Network}
\label{sec: Graph Convolutional Network}
Graph Convolutional Network (GCN) \cite{kipf2017semisupervised} is introduced to learn the knowledge between the neighbourhood information propagation labels of a node. The GCN can learn more complex representations by stacking multiple graph convolutional layers. Each GCN layer can be calculated using the following formula:
\begin{equation}
\label{eq: GCN}
{\boldsymbol{H}}^{l+1}=\mathcal{F}(\hat{\boldsymbol{A}}{\boldsymbol{H}}^{l}{\boldsymbol{W}}^{l})
\end{equation}
where ${\boldsymbol{W}}^{l}\in {\mathbb{R}}^{d\times d’}$ represents a transformation matrix to be learned ($d$ and $d'$ denote the dimensions of node features); ${\hat{\boldsymbol{A}}}^{l}\in {\mathbb{R}}^{n\times n}$ indicates the normalised correlation matrix $\boldsymbol{A}$ ($n$ is the number of nodes); $\mathcal{F}(\cdot)$ represents a non-linear function, which is acted by LeakyReLU in our experiments; ${\boldsymbol{H}}^{l}\in {\mathbb{R}}^{n\times d}$ and ${\boldsymbol{H}}^{l+1}\in {\mathbb{R}}^{n\times d’}$ are the node representation matrices for the $l$-th and $(l+1)$-th layers, respectively, with each row corresponding to a node in the graph and each column corresponding to the features of this node. In our method, GCN is employed to learn hierarchical knowledge between labels/classes in the hierarchical knowledge learning module.

%  可添加可删除
% And ${X}_{w}[k]$ represents the spectrum value of the original signal sequence ${x[n]}$ at the discrete frequency point ${w_k} = {{2\pi k} \mathord{\left/{\vphantom {{2\pi k} N}} \right.\kern-\nulldelimiterspace} N}$. Specifically, ${X}_{w}[k]$ is equivalent to applying an equal interval sampling to the spectrum ${X}_{w}[{e^{jw}}]$ in the range $\left[ {0,2\pi } \right)$ with a sampling interval $\Delta w = {{2\pi } \mathord{\left/{\vphantom {{2\pi } N}} \right.\kern-\nulldelimiterspace} N}$.
% GCN updates each node by considering neighbourhood information of the node, which better captures the information in the graph structure. 
% In contrast to previous structured cavitation intensity recognition focusing on complex network design, HKG directly formulates it as a multi-label classification task. Furthermore, GCN-based hierarchical knowledge is embedded in deep features to achieve HKG cavitation intensity recognition.

\section{Method}
\label{sec Method}
\subsection{Architecture Overview}
\label{subsec: Architecture Overview}
The key idea of \underline{h}ierarchical \underline{k}nowledge-\underline{g}uided fault intensity diagnosis (HKG) is to exploit the inherent hierarchical knowledge of the cavitation classes learned by graph convolutional network (GCN) to guide/constrain the deep features learned by CNN. This integrated approach can capture subtle nuances in acoustic signals and helps to improve the quantitative relationship between signal features and fault intensity. In our design, one stream is employed for fault feature learning and the other stream is used for hierarchical knowledge learning. The overall architecture of our model is depicted in Figure \ref{fig: framework} (see appendix for HKG algorithm).
\begin{figure*}
    \centering
    \includegraphics[width=0.95\textwidth,height=58mm]{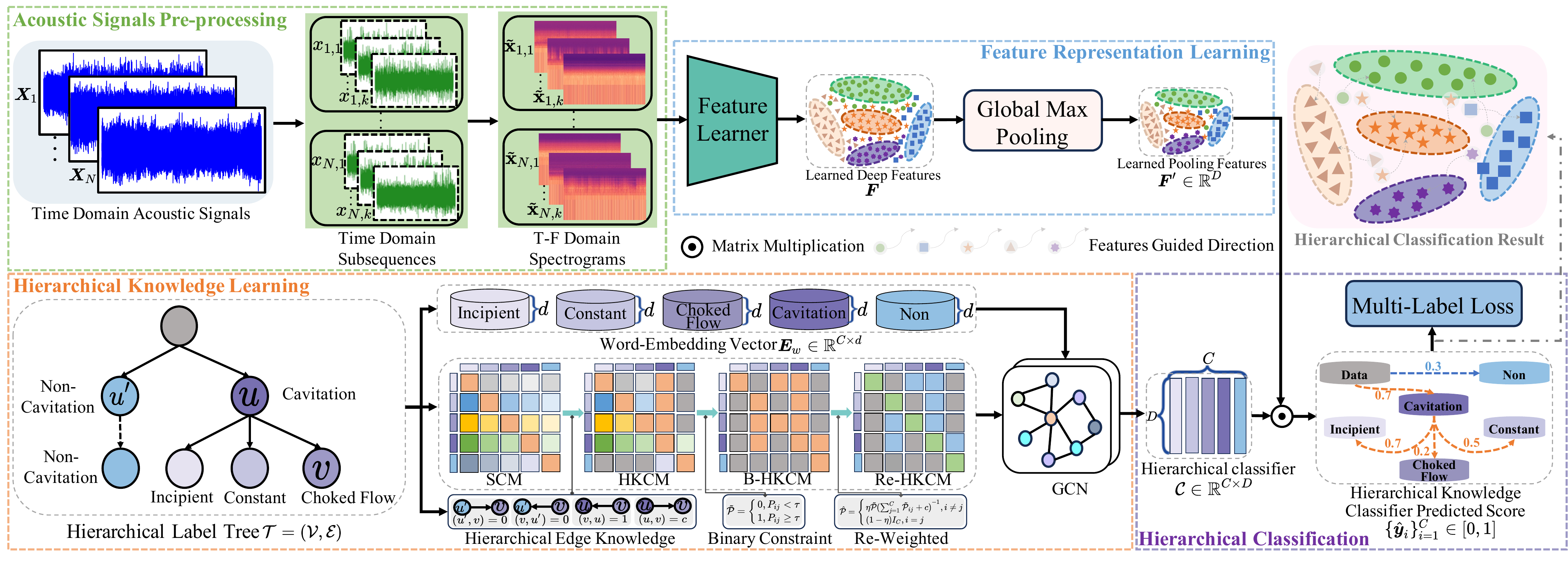}
    \caption{Overall framework of the HKG. The T-F domain spectrograms are fed into feature representation learning module for extracting deep features ${\boldsymbol{F}'}$. Meanwhile, the target classes are used to generate word embeddings ${\boldsymbol{E}_w}$ and re-weighted hierarchical knowledge correlation matrix $\tilde{\boldsymbol{A}}$, which are input to a GCNs to generate a set of interdependent global hierarchical knowledge classifiers $\mathcal{C}$. Finally, the deep features ${\boldsymbol{F}'}$ and the hierarchical knowledge classifier $\mathcal{C}$ are obtained predicted scores.}
    \label{fig: framework}
\end{figure*}

Given a signal dataset ${\boldsymbol{X}} = \{ {X_i},i = 1,2, \ldots ,N\}  \subseteq {\mathbb{R}^{M \times N}}$ and the corresponding label ${\boldsymbol{L}} \in {\mathbb{R}^{C \times N}}$ with $N$ streams, $M$ measurements for each stream and $C$ fault classes. For the fault feature learning stream, the signal dataset ${\boldsymbol{X}}$ is input to the signal pre-processing module and ${\boldsymbol{\tilde X}} \in {\mathbb{R}^{T \times F \times 3}}$ is the output after the sliding window (SW) and STFT operations as follows:
\begin{equation}
\label{eq: acoustic signals pre-processing}
\begin{array}{l}
{\boldsymbol{\tilde X}} = 10 \times {\log _{10}}(\frac{{\left| {\mathrm{STFT}(\mathrm{SW}(\boldsymbol{X}))} \right|}}{{\max (\left| {\mathrm{STFT}(\mathrm{SW}(\boldsymbol{X}))} \right|)}})\\
\;\;\;: = 10 \times {\log _{10}}(\frac{{\left| {{\boldsymbol{X}_{sw}}[n,m]} \right|}}{{\max (\left| {{\boldsymbol{X}_{sw}}[n,m]} \right|)}})
\end{array}
\end{equation}
where ${{\boldsymbol{X}}_{sw}}[n,m]$ indicates the $n$-th row and $m$-th column elements of the results matrix of STFT and $\left| {{\boldsymbol{X}_{sw}}[n,m]} \right|$ denotes the elements of the amplitude spectrum matrix. Then, ${\boldsymbol{\tilde X}}$ is fed into the feature representation learning module (FRL) to produce learned pooling features ${\boldsymbol{F}'} \in {{\mathbb{R}}^D}$ with $D$ denoting the dimension of the T-F domain spectrogram. For hierarchical knowledge learning stream, the label set ${\boldsymbol{L}}$ is first one-hot encoded and then hierarchical label tree ${\mathcal{T}} \in {{\mathbb{R}}^{C \times H}}$ are created by manual or unsupervised learning, where $H$ is the height of the tree. Next, the hierarchical label tree ${\mathcal{T}} \in {{\mathbb{R}}^{C \times H}}$ is used as input to the hierarchical knowledge learning module (HKL) to generate a hierarchical knowledge classifier ${\mathcal{C}} \in {\mathbb{R}^{C \times D}}$. Finally, the hierarchical knowledge guided predicted score $\hat{\boldsymbol{Y}} \in {{\mathbb{R}}^C}$ can be computed by:
\begin{equation}
\label{eq: hierarchical knowledge guided features}
\hat{\boldsymbol{Y}}=\mathrm{FRL}({\boldsymbol{\tilde X}})\cdot {(\mathrm{HKL}(\mathcal{T}))}^{\top} :=\boldsymbol{F}’ \cdot \mathcal{C}^\top 
\end{equation}
where $\cdot$ represents matrix multiplication operation and $\mathcal{C}^\top$ denotes the transpose matrix of $\mathcal{C}$. Through the described process, our HKG exploits hierarchical knowledge of target states to effectively constrain the extracted deep features. The effect of hierarchical classification is shown in the top right corner of Figure \ref{fig: framework}.

\subsection{Feature Representation Learning}
\label{subsec: Feature Representation Learning}
The feature representation learning uses deep learning methods to extract deep features from T-F domain spectrograms and the extracted features are acted on hierarchical classifiers. 

In our experiments, we employ various deep learning methods as feature learners, including convolutional neural networks (ResNet \cite{he2016deep}, VGG \cite{simonyan2014very} and DenseNet \cite{huang2017densely}), lightweight neural networks (MobileNetv2/v3 \cite{sandler2018mobilenetv2,howard2019searching} and ShuffleNetv2 \cite{ma2018shufflenet}) and transformers (ViT \cite{dosovitskiy2020image} and Swin \cite{liu2021swin}). Given a T-F domain spectrogram ${\boldsymbol{\tilde X}}$ as an input to the feature learner ${{\mathcal F}_{\mathrm{FL}}}$, we can obtain a learned deep feature $\boldsymbol{F}$. Next, we perform a global max pooling layer ${{\mathcal F}_{\mathrm{GMP}}}$ to obtain the learned pooling feature $\boldsymbol{F}'\in {{\mathbb{R}}^D}$. The mathematical flow of feature representation learning is described as follows:
\begin{equation}
\label{eq: feature representation learning}
\boldsymbol{F}' = {{\mathcal F}_{\mathrm{GMP}}}({{\mathcal F}_{\mathrm{FL}}}({{\boldsymbol{\tilde X}}},\boldsymbol{\theta}_\mathrm{FL} )): = {{\mathcal F}_{\mathrm{GMP}}}(\boldsymbol{F})
\end{equation}
where $\boldsymbol{\theta}_\mathrm{FL}$ is the parameter of the feature learner and $D = T \times F \times {C_h}$ denotes the dimension of the T-F domain spectrogram ($T$, $F$ and ${C_h}$ are the width, height and number of channels of the spectrogram.) 

\subsection{Hierarchical Knowledge Learning}
\label{subsec: Hierarchical Knowledge Learning}
The core of GCN is to propagate information between nodes based on the correlation matrix. Therefore, how to construct the correlation matrix is a key issue in GCN. In this paper, we construct a correlation matrix in a data-driven and structured knowledge way. In other words, we define the correlation between classes by mining the co-occurrence patterns of classes from the dataset and correcting the correlation based on the inherent hierarchical knowledge in the hierarchical label tree.

\noindent\textbf{Class Dependency.} We represent class dependencies in terms of conditional probability, i.e., ${A_{i \to j}} = P({L_j}|{L_i})$ denotes the importance of class ${L_{i \in C}}$ for class ${L_{j\in C}}$ and ${A_{j \to i}} = P({L_i}|{L_j})$ indicates the importance of class ${L_{j\in C}}$ for class ${L_{i \in C}}$.

\noindent\textbf{Statistical Correlation Matrix (SCM).} To construct the statistical correlation matrix, we first count the number of each class in the dataset to obtain $\boldsymbol{S} = \{ {S_i}\} _{i = 1}^C$. Then, we calculate the count of co-occurrences of class $L_i$ and class $L_j$, i.e. $S(i,j) = S(j,i) = {S_i} + {S_j}(i \ne j)$. Finally, the SCM $\boldsymbol{A}\in \mathbb{R}^{C\times C}$ can be obtained as follows:
\begin{equation}
\label{eq: SCM}
\boldsymbol{A}=\left [{A}_{ij}\right]=\begin{cases}
 {A}_{ij}=\frac{{S}_{i}}{{S}_{i}+{S}_{j}},\,\,\,\,\,i\neq j \\ 
 {A}_{ii}=\frac{{S}_{i}}{{S}_{i}}=\frac{{S}_{j}}{{S}_{j}},i=j
\end{cases}
\end{equation}
According to Equation \ref{eq: SCM}, we can get ${A_{i \to j}} = P({L_j}|{L_i}) = {{{S_i}} \mathord{\left/{\vphantom {{{S_i}} {({S_i} + {S_j})}}} \right.\kern-\nulldelimiterspace} {({S_i} + {S_j})}}$ and 
${A_{j \to i}} = P({L_i}|{L_j}) = {{{S_j}} \mathord{\left/{\vphantom {{{S_i}} {({S_i} + {S_j})}}} \right.\kern-\nulldelimiterspace} {({S_i} + {S_j})}}$. Therefore, the SCM is an asymmetrical matrix with a diagonal of 1 since ${A_{i \to j}} = P({L_j}|{L_i}) \ne {A_{j \to i}} = P({L_i}|{L_j})$ and ${A_{ii}} = {{{S_i}} \mathord{\left/{\vphantom {{{S_i}} {{S_i}}}} \right.\kern-\nulldelimiterspace} {{S_i}}}$. 

\noindent\textbf{Hierarchical Knowledge Correlation Matrix (HKCM).} The basic SCM does not consider hierarchical knowledge among classes. Therefore, we embed hierarchical edge knowledge based on hierarchical label tree ${\mathcal{T}} = ({\mathcal{V}},{\mathcal{E}})$ into the SCM to achieve the potential dependencies between classes are considered. Each node $v \in {\mathcal{V}}$ denotes a target class/concept and each edge $(u,v) \in {\mathcal{E}}$ encodes the decomposition relationship between class $u \in {\mathcal{V}}$ and class $v \in {\mathcal{V}}$, where the parent node $u$ is a broader conceptual superclass of the child node $v$, such as $(u,v) = (\mathrm{cavitation},\mathrm{incipient\;cavitation})$. In addition, node ${u'} \in {\mathcal{V}}$ and node $u$ is the same hierarchical classes and are siblings of each other, i.e., ${u'}$ is a cousin node of $v$, such as $(u',v) = (\mathrm{non\;cavitation},\mathrm{incipient\;cavitation})$. We also assume $(v,v)$, i.e., each class is not only a subclass of itself, but also a superclass of itself. Therefore, we have the following hierarchical edge knowledge constraints:
\begin{equation}
\label{eq: hierarchical knowledge}
\begin{cases}
{{\mathcal{E}}_{v \to u}} = (v,u) = 1\\
{{\mathcal{E}}_{v \to u'}} = (v,u') = 0\\
{{\mathcal{E}}_{u' \to v}} = (u',v) = 0
\end{cases}   
\end{equation}
Next, we create a transition matrix $\Phi \in \mathbb{R}^{C\times C}$ with a diagonal of 1, the elements corresponding to constraints ${{\mathcal{E}}_{v \to u'}}$ and ${{\mathcal{E}}_{u' \to v}}$ as 0, the element corresponding to the constraint ${{\mathcal{E}}_{v \to u}}$ as $1/A_{ij}(i \ne j)$ and the remaining elements as 1. Then, the HKCM $\check {\boldsymbol{A}} \in \mathbb{R}^{C\times C}$ can be acquired as follows:
\begin{equation}
\label{eq: HKCM}
\check {\boldsymbol{A}} = \boldsymbol{A} \odot \Phi 
\end{equation}
where $\odot$ is the Hadamard product. Since the $\boldsymbol{A}$ is an asymmetrical matrix with diagonal 1 and $\left| \boldsymbol{A} \right| \ne 0$, $\boldsymbol{A}{\boldsymbol{A}^{ - 1}} = \boldsymbol{I}$ ($\boldsymbol{I} \in \mathbb{R}^{C\times C}$ is a unit matrix), so the $\boldsymbol{A}$ exists an inverse matrix ${\boldsymbol{A}^{ - 1}}$. Furthermore, we have $\Phi  = {\boldsymbol{A}^{ - 1}}\check {\boldsymbol{A}}$, which indicates that hierarchical edge knowledge is interpretable and equivalent to conventional matrix transformations.

\noindent\textbf{Binary HKCM (B-HKCM).} The HKCM may face two drawbacks. Firstly, the pattern of co-occurrence between class and other classes may suffer from a long-tailed distribution \cite{wang2020multi}. Secondly, the absolute number of training and testing datasets may not be exactly equivalent. Therefore, we suggest to binarise the HKCM $\check {\boldsymbol{A}}$ to filter the edge noise. The B-HKCM $\hat{\boldsymbol{A}}\in \mathbb{R}^{C\times C}$ is derived as follows:
\begin{equation}
\label{eq: B-HKCM}
\hat{\boldsymbol{A}}=\begin{cases}
 0, \,\,\,\, \mathrm{if} \, {\check{\boldsymbol{A}}}_{ij}<\tau \\ 
 1, \,\,\,\, \mathrm{if} \, {\check{\boldsymbol{A}}}_{ij}\geq \tau
\end{cases}
\end{equation}
where $\tau  \in \left( {0,1} \right]$ is the threshold and ${\check{\boldsymbol{A}}}_{ij}$ denotes the element of the $i$-th row and $j$-th column of $\check {\boldsymbol{A}}$.

\noindent\textbf{Re-weighted HKCM (Re-HKCM).} The B-HKCM only considers the connection relationship between nodes and ignores the strength information of the connections. This can cause features to blend too evenly across the graph as the GCN propagates information, blurring the differences between nodes, i.e., over-smoothing \cite{liu2023enhancing, chen2019multi}. Our proposed re-weighting scheme considers the connection strength of nodes and introduces more detailed information to retain more differences in node features. This can more effectively distinguish between nodes of various clusters and improve the discriminative ability of the model. Specifically, the re-weighting scheme is as follows:
\begin{equation}
\label{eq: Re-HKCM}
\tilde{\boldsymbol{A}}=\begin{cases}
 \frac{\eta \hat{\boldsymbol{A}}}{\sum_{j=1}^{C}{\hat{\boldsymbol{A}}}_{ij}+c}, \,\,\,\,i\neq j\\ 
 (1-\eta ){\boldsymbol{I}},\,\,\,\,\,\,\,\,\,i=j
\end{cases}
\end{equation}
where $\tilde{\boldsymbol{A}} \in \mathbb{R}^{C\times C}$ is Re-HKCM, ${\hat{\boldsymbol{A}}}_{ij}$ denotes the element of the $i$-th row and $j$-th column of $\hat {\boldsymbol{A}}$, $\eta  \in \left[ {0,1} \right]$ is the scaling factor to adjust the matrix elements, $c$ denotes the smoothing factor to avoid infinitely large values and $\boldsymbol{I}$ represents the matrix of ${C\times C}$. When $\eta  \to 1$, the neighborhood information is emphasized. When $\eta  \to 0$, the node's features are considered.

\noindent\textbf{Word Embedding.} The word embedding represents the semantic relationships between classes by mapping words into a continuous vector space, which better models associations of classes. In our model, the hierarchical label tree ${\mathcal{T}} \in {{\mathbb{R}}^{C \times H}}$ is used as an input to the global vectors for word representation (GloVe) \cite{pennington2014glove} to a generate word embedding vector ${\boldsymbol{E}_w} \in \mathbb{R}^{C\times d}$ ($d$ is the dimension of the class-level word embedding), which is used as one of the inputs to GCN. 

Based on the above description, the global hierarchical knowledge classifier ${\mathcal{C}} \in {\mathbb{R}^{C \times D}}$ generated by the hierarchical knowledge learning module is as follows:
\begin{equation}
\label{eq: HKL}
\begin{array}{l} {\mathcal{C}}=\mathcal{F}_{\mathrm{GCN}}((\mathrm{GloVe}(\mathcal{T}),\tilde{\boldsymbol{A}}),{\boldsymbol{\theta}}_{\mathrm{GCN}}))\\
\,\,\,\,\,\,:={\mathcal{F}_{\mathrm{GCN}}((\boldsymbol{E}_{w},\tilde{\boldsymbol{A}}),{\boldsymbol{\theta}}_{\mathrm{GCN}})} \end{array}
\end{equation}
where $\boldsymbol{\theta}_\mathrm{GCN}$ is the parameter of the GCN. 
% For hierarchical knowledge learning, two GCN layers are stacked. For the first layer, $\boldsymbol{E}_{w}$ and $\tilde{\boldsymbol{A}}$ are used for inputs. For the last layer, the output is ${\mathcal{C}}$.

\subsection{Hierarchical Classification}
\label{subsec: Hierarchical Classification}
\noindent\textbf{Hierarchical Knowledge Classifier Predicted Score.} An inter-dependent global hierarchical knowledge classifier, i.e., ${\mathcal{C}} \in {\mathbb{R}^{C \times D}}$ is learned from label representations based on the mapping function of GCN, where $C$ is the number of classes and $D$ indicates the dimension of the learned deep pooling features $\boldsymbol{F}' $. We employ the learned global  hierarchical knowledge classifier ${\mathcal{C}}$ on the learned deep pooling features $\boldsymbol{F}' $ to obtain the prediction scores $\hat{\boldsymbol{Y}}$, as follows:
\begin{equation}
\label{eq: hierarchical knowledge classifier}
\hat{\boldsymbol{Y}}=\{ {{\hat{\boldsymbol{y}}_i}}\} _{i = 1}^C=\boldsymbol{F}’ \cdot \mathcal{C}^\top 
\end{equation}
where ${{\hat{\boldsymbol{y}}_i}} \in [0,1]$ is the predicted score for each class.

\noindent\textbf{Objective Function.} In contrast to previous structured cavitation intensity diagnoses focusing on complex network design, HKG directly formulates it as a multi-label classification task. Therefore, our method uses the traditional multi-label classification loss as our model's objective function, as follows:
\begin{equation}
\label{eq: objective function}
\mathcal{L}=-\frac{1}{C}\sum_{i=1}^{C}{\boldsymbol{l}}_{i}\log(\frac{1}{1+{e}^{-{\hat{\boldsymbol{y}}}_{i}}})+(1-{\boldsymbol{l}}_{i})\log(\frac{{e}^{-{\hat{\boldsymbol{y}}}_{i}}}{1+{e}^{-{\hat{\boldsymbol{y}}}_{i}}})
\end{equation}
where $C$ is the number of classes, $\boldsymbol{l}$ denotes the ground truth label of a T-F domain sub-spectrogram and $\hat{\boldsymbol{y}}$ indicates the prediction score of our model. The proposed HKG is summarized in the Algorithm \ref{algo: HKG Algorithm}.
\begin{algorithm}[htbp]
    \caption{HKG-based Fault Intensity Diagnosis}
    \label{algo: HKG Algorithm}
    \begin{algorithmic}
    \Require original signal dataset ${\boldsymbol{X}}\subseteq {\mathbb{R}^{M \times N}}$, the corresponding label ${\boldsymbol{L}} \in {\mathbb{R}^{C \times N}}$, threshold $\tau  \in \left( {0,1} \right]$ and scaling factor $\eta  \in \left[ {0,1} \right]$ 
    \Ensure hierarchical knowledge classifier predicted score ${{\hat{\boldsymbol{y}}_i}}$
    \For{epoch $= 0,1,\ldots,N$}
    \State \textbf{Acoustic Signals Pre-processing:}
    \State ${\boldsymbol{X}} = \{ {X_i},i = 1,2, \ldots ,N\} \Rightarrow {\boldsymbol{\tilde X}} \in {\mathbb{R}^{T \times F \times 3}}$  \hfill$\rhd$ Equation \ref{eq: acoustic signals pre-processing}
    \State \textbf{Feature Representation Learning:}
    \State ${{\mathcal F}_{\mathrm{GMP}}}({{\mathcal F}_{\mathrm{FL}}}({{\boldsymbol{\tilde X}}},\boldsymbol{\theta}_\mathrm{FL} )) \Rightarrow \boldsymbol{F}'\in {{\mathbb{R}}^D}$ \hfill$\rhd$ e.g. CNNs, Transformer
    \State \textbf{Hierarchical Knowledge Learning:}
    \State Create hierarchical label tree:
    \State ${\boldsymbol{L}} \in {\mathbb{R}^{C \times N}} \Rightarrow {\mathcal{T}} \in {{\mathbb{R}}^{C \times H}}$ \hfill$\rhd$ Manual, Unsupervised Learning
    \State Build correlation matrix:
    \State ${\mathcal{T}} = ({\mathcal{V}},{\mathcal{E}}) \Rightarrow \boldsymbol{A}\in \mathbb{R}^{C\times C}$ \hfill$\rhd$ Equation \ref{eq: SCM}
    \State $\boldsymbol{A}\in \mathbb{R}^{C\times C} \stackrel{\varepsilon} {\Rightarrow} \check {\boldsymbol{A}} \in \mathbb{R}^{C\times C}$  \hfill$\rhd$ Equations \ref{eq: hierarchical knowledge} and \ref{eq: HKCM}
    \State $\check {\boldsymbol{A}} \in \mathbb{R}^{C\times C} \stackrel{\tau} {\Rightarrow} \hat{\boldsymbol{A}}\in \mathbb{R}^{C\times C}$ \hfill$\rhd$ Equation \ref{eq: B-HKCM}
    \State $\hat{\boldsymbol{A}}\in \mathbb{R}^{C\times C} \stackrel{\eta} {\Rightarrow} \tilde{\boldsymbol{A}} \in \mathbb{R}^{C\times C}$ \hfill$\rhd$ Equation \ref{eq: Re-HKCM}
    \State Generate word-embedding vector:
    \State $\mathrm{Word2Vec}(\mathcal{T}) \Rightarrow {\boldsymbol{E}_w} \in \mathbb{R}^{C\times d}$ \hfill$\rhd$ e.g. GloVe, FastText
    \State Hierarchical knowledge classifier:
    \State ${\mathcal{F}_{\mathrm{GCN}}((\boldsymbol{E}_{w},\tilde{\boldsymbol{A}}),{\boldsymbol{\theta}}_{\mathrm{GCN}})} \Rightarrow {\mathcal{C}} \in {\mathbb{R}^{C \times D}}$ \hfill$\rhd$ Equation \ref{eq: HKL}
    \State \textbf{Hierarchical Classification:}
    \State Obtain prediction score $\hat{\boldsymbol{Y}}$:
    \State $\boldsymbol{F}’ \cdot \mathcal{C}^\top \Rightarrow \hat{\boldsymbol{Y}}=\{ {{\hat{\boldsymbol{y}}_i}}\} _{i = 1}^C$ \hfill$\rhd$ Equation \ref{eq: hierarchical knowledge classifier}
    \State Update parameters of HKG by minimizing $\mathcal{L}$:
    \State min $-\frac{1}{C}\sum_{i=1}^{C}{\boldsymbol{l}}_{i}\log(\frac{1}{1+{e}^{-{\hat{\boldsymbol{y}}}_{i}}})+(1-{\boldsymbol{l}}_{i})\log(\frac{{e}^{-{\hat{\boldsymbol{y}}}_{i}}}{1+{e}^{-{\hat{\boldsymbol{y}}}_{i}}})$
    \State Save parameters $\boldsymbol{\theta}_\mathrm{FL}$ and ${\boldsymbol{\theta}}_{\mathrm{GCN}}$ of HKG in current epoch.
    \EndFor
	\end{algorithmic}
\end{algorithm}

\section{Experiments}
\label{sec: Experiments}
We conduct extensive experiments on four different datasets (three cavitation datasets are provided by SAMSON AG and one public dataset) to validate the effectiveness and scalability of the proposed HKG. And detailed analysis of the performance and its comparison with state-of-the-art methods are also reported. In addition, we provide extensive discussions and ablation analysis to demonstrate the stability and significance of the proposed HKG.

\subsection{Evaluation Metrics}
\label{subsec: Evaluation Metrics}
To achieve hierarchical knowledge-guided fault intensity diagnosis, we technically treat it as a multi-label task. We still need to use the same evaluation metrics as traditional fault intensity diagnosis from the from the practical application for comparability and interpretability. Therefore, we select Accuracy (Acc), Precision (Pre), Recall (Rec) and F1-score (F1) as evaluation metrics following previous studies \cite{sha2022regional,fedorishin2022large}. The threshold is determined by the Youden index in ROC-AUC (more details see Appendix).

\subsection{Implementation Details}
\label{subsec: Implementation Details}
The HKG contains a feature learning stream and a hierarchical knowledge learning stream. For the feature learning stream, ResNet \cite{he2016deep}, VGG \cite{simonyan2014very}, DeneseNet \cite{huang2017densely}, MobNetv2/3 \cite{sandler2018mobilenetv2,howard2019searching}, ShuNetv2 \cite{ma2018shufflenet}, ViT \cite{dosovitskiy2020image} and Swin \cite{liu2021swin} are used as feature learners, respectively. And the input T-F domain spectrogram is flipped horizontally, flipped vertically, rotated with \SI{180}{\degree} and resized to 512 $\times$ 512. For the hierarchical knowledge learning stream, 2 layers of GCN are used with output dimensions of 512 and 1024, respectively (not specified). In addition, LeakyReLU non-linear function is applied across the two GCN layers. For word embeddings, we use a 300-dim GloVe trained on the Wikipedia corpus (the default word embedding method). For the correlation matrix, the parameters $\tau$ and $\eta$ in Equation \ref{eq: B-HKCM} and Equation \ref{eq: Re-HKCM} are set to 0.3 and 0.4, respectively. During training, HKG uses SGD as the optimizer with a momentum of 0.9 and a weight decay of ${10^{ - 4}}$. The initial learning rate of the optimizer is set to 0.01 and dynamically adjusted by the Plateau Scheduler. The HKG is trained for 100 epochs with a batch size of 16. Our source code is released at \textcolor{orange}{\url{https://github.com/CavitationDetection/HKG}}.

\subsection{Datasets}
\label{subsec: Datasets}
\noindent\textbf{Cavitation Datasets.} This dataset is provided by SAMSON AG and contains three sub-datasets, called \textbf{Cavitation-Short}, \textbf{Cavitation-Long} and \textbf{Cavitation-Noise} with real noise, respectively. The cavitation acoustic signals are collected from different valves with different upstream pressures and different valve opening rates in a professional environment. The cavitation includes incipient cavitation, constant cavitation and choked flow cavitation. The non-cavitation contains turbulent flow and no flow. Cavitation-Short has a total of 356 acoustic signals and each acoustic signal with time duration of \SI{3}{s}. Cavitation-Long and Cavitation-Noise have 806 and 160 acoustic signals and each of them with time duration of \SI{25}{s}, respectively. The sampling rate of all recording acoustic signals is \SI{1562.5}{kHz} (more details see Appendix).

\noindent\textbf{PUB Dataset.} This dataset \cite{lessmeier2016condition} contains bearing vibration signals provided by Paderborn University. The PUB contains three bearing states named Inner Ring (IR) Damage, Outer Ring (OR) Damage and Healthy, with each fault state comprising artificial damage and real damage. The vibration signals of the PUB are collected at a sampling frequency of \SI{64}{kHz} with a time duration of \SI{4}{s}. The PUB is organized into three hierarchies based on damage areas and damage levels of the bearing: baring diagnosis (Hierarchy I), damage type diagnosis (Hierarchy II) and IR/OR intensity diagnosis (Hierarchy III-IR/III-OR). In our experiments, $80\%$ of data is allocated to model training and $20\%$ of data is used for model testing (more details see Appendix). 

\subsection{Experimental Results}
\label{subsec: Experimental Results}
\begin{figure}[htbp]
\centering
\subfigure[ResNet34]{
\begin{minipage}[t]{0.5\linewidth}
\centering
\includegraphics[width=\textwidth,height=35mm]{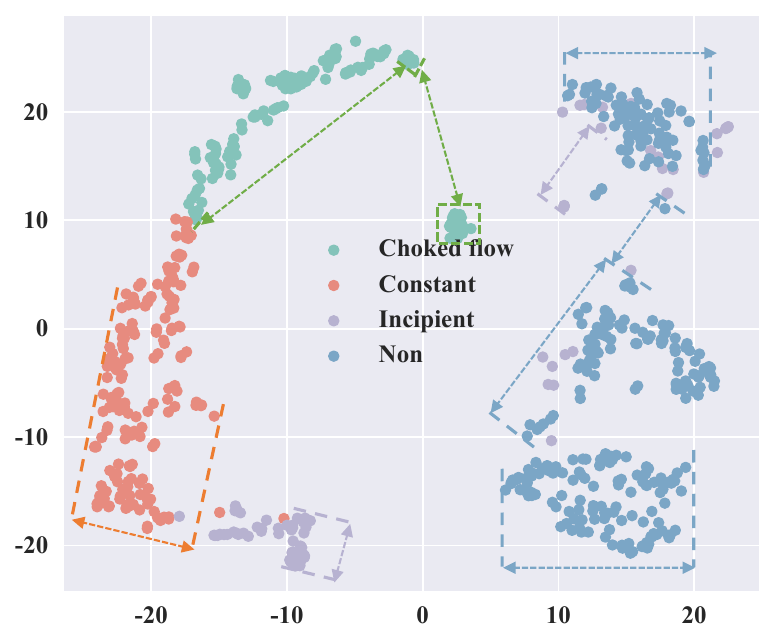}
%\caption{fig1}
\end{minipage}%
}%
\subfigure[HKG+ResNet34]{
\begin{minipage}[t]{0.5\linewidth}
\centering
\includegraphics[width=\textwidth,height=35mm]{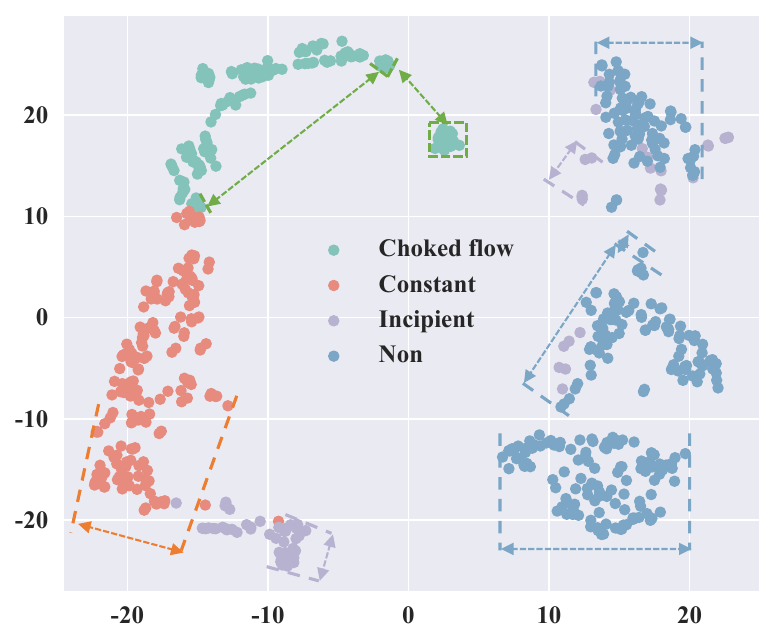}
%\caption{fig1}
\end{minipage}%
}%
\centering
\caption{Visualisation of the learned deep feature distribution of vanillia ResNet34 and HKG+ResNet34 via t-SNE \cite{van2008visualizing} on Cavitation-Short.}
\label{fig: examples features}
\end{figure}

\noindent\textbf{Results on Cavitation datasets.} Figure \ref{fig: examples features} shows examples of cavitation deep feature distributions with and without the guidance of hierarchical knowledge. It indicates that hierarchical knowledge can better constrain and guide cavitation deep features. As seen in Table \ref{tab: Different Cavitation Results}, HKG shows superior results under different representation learning methods across different cavitation datasets. The proposed HKG achieves the best accuracy of \textbf{89.71}$\%$, \textbf{93.18}$\%$ and \textbf{99.63}$\%$ in three cavitation datasets, respectively. It shows that our proposed HKG can be applied to any representation learning framework.

Detailedly, (1) \noindent\textbf{Cavitation-Short:} The accuracy of HKG is above \textbf{85}$\%$ for each specific backbone, which outperforms the corresponding baseline. The HKG improves the accuracy of ResNet34 by \textbf{1.14}$\%$. Specifically, the HKG achieves \textbf{92.35}$\%$ precision on the Swin-B. (2) \noindent\textbf{Cavitation-Long:} The accuracy of HKG under ResNet34, DenseNet169, ViT-B (32) and Swin-B (4-12) as backbones all exceed \textbf{92}$\%$, which improves \textbf{0.56}$\%$, \textbf{0.61}$\%$, \textbf{0.44}$\%$ and \textbf{0.3}$\%$ over the corresponding backbone, respectively. The HKG achieves the best precision, recall and F1-score on Swin-B with \textbf{89.58}$\%$, \textbf{93.40}$\%$ and \textbf{91.27}$\%$, respectively. (3) \noindent\textbf{Cavitation-Noise:} Our method obtains perfect performance in all backbones, which is better than all baselines.The  HKG obtains an accuracy of \textbf{99.25}$\%$, \textbf{99.06}$\%$, \textbf{99.19}$\%$ and \textbf{99.63}$\%$ on backbone of ResNet18, ResNet34, DeneseNet169 and Swin-B, respectively. And it has an improvement of \textbf{0.31}$\%$, \textbf{0.25}$\%$, \textbf{0.5}$\%$ and \textbf{0.44}$\%$ over the corresponding backbone. In addition, the best precision, recall and F1-score obtained by HKG+Swin-B are \textbf{99.63}$\%$, \textbf{99.63}$\%$ and \textbf{99.62}$\%$, respectively. The reasons why our method gets superior results are as follows. First, although this dataset has real background noise compared to other cavitation datasets, the time-frequency transform can filter most of the noise. Second, this dataset is obtained with only one operating of valve stroke and upstream pressure. Third, each class of data is balanced for this dataset (see Appendix).
\begin{table}[htbp]
\caption{Results of different evaluation metrics on three real-world cavitation datasets. We compare different representation learning methods including CNNs, LNNs and Transformers. According to our previous research \cite{sha2022multi}, the sliding window with a window size of 466944 and a step size of 466944.}
\label{tab: Different Cavitation Results}
\scriptsize
\setlength{\tabcolsep}{0.1mm}{
\begin{tabular}{l|c|cccc|cccc|cccc}
\toprule
\multicolumn{2}{c}{Dataset}      & \multicolumn{4}{c}{Cavitation-Short}  & \multicolumn{4}{c}{Cavitation-Long}  & \multicolumn{4}{c}{Cavitation-Noise}    \\ 
\midrule
\multicolumn{1}{c|}{Methods}    & \multicolumn{1}{c|}{\begin{tabular}[c]{@{}c@{}}Image \\ size\end{tabular}} & \multicolumn{1}{c}{Acc} & \multicolumn{1}{c}{Pre} & \multicolumn{1}{c}{Rec} & \multicolumn{1}{c|}{F1} 
                                & \multicolumn{1}{c}{Acc} & \multicolumn{1}{c}{Pre} & \multicolumn{1}{c}{Rec} & \multicolumn{1}{c|}{F1} 
                                & \multicolumn{1}{c}{Acc} & \multicolumn{1}{c}{Pre} & \multicolumn{1}{c}{Rec} & \multicolumn{1}{c}{F1} \\ 
\midrule
ResNet18 \cite{he2016deep}                      &${256^2}$       &87.57  &90.25  &77.16  &77.95     &90.83  &86.31  &90.75  &88.21      &98.94  &98.94  &98.94  &98.94              \\
ResNet34 \cite{he2016deep}                      &${256^2}$       &88.57  &\underline{91.47}  &76.22  &74.00     &91.88  &87.60  &91.86  &89.42      &98.81  &98.82  &98.81  &98.81                         \\
ResNet50 \cite{he2016deep}                      &${256^2}$       &84.43  &86.97  &71.70  &66.83     &91.45  &87.09  &91.25  &88.89      &98.19  &98.20  &98.19  &98.19                         \\
ResNet101 \cite{he2016deep}                     &${256^2}$       &82.14  &59.88  &68.75  &63.73     &89.49  &84.54  &89.43  &86.56      &97.56  &97.57  &97.56  &97.56                         \\
VGG11 \cite{simonyan2014very}                   &${256^2}$       &84.29  &62.31  &70.10  &65.95     &80.78  &74.46  &81.11  &76.69      &97.81  &97.82  &97.81  &97.81                         \\
VGG13 \cite{simonyan2014very}                   &${256^2}$       &80.71  &58.96  &66.85  &62.60     &81.56  &75.21  &82.00  &77.50      &98.19  &98.19  &98.19  &98.19                         \\
VGG16 \cite{simonyan2014very}                   &${256^2}$       &79.86  &57.77  &66.53  &61.25     &83.38  &77.30  &83.14  &79.44      &98.44  &98.44  &98.44  &98.44                         \\
VGG19 \cite{simonyan2014very}                   &${256^2}$       &73.29  &64.40  &59.76  &59.03     &82.80  &76.52  &82.42  &78.63      &98.69  &98.69  &98.69  &98.69                         \\
DenseNet121 \cite{huang2017densely}             &${256^2}$       &84.86  &87.36  &72.01  &67.21     &91.16  &86.73  &91.24  &88.66      &97.94  &97.94  &97.94  &97.94                         \\
DenseNet161 \cite{huang2017densely}             &${256^2}$       &85.00  &89.49  &71.96  &71.14     &91.84  &87.84  &92.07  &89.67      &98.25  &98.25  &98.25  &98.25                         \\
DenseNet169 \cite{huang2017densely}             &${256^2}$       &87.00  &64.39  &73.04  &68.28     &92.08  &87.78  &92.14  &89.65      &98.69  &98.69  &98.69  &98.69                         \\
DenseNet201 \cite{huang2017densely}             &${256^2}$       &84.14  &86.59  &72.13  &69.37     &90.40  &85.66  &90.79  &87.76      &98.56  &98.57  &98.56  &98.56                         \\
\rowcolor[HTML]{EFEFEF}\textbf{HKG-ResNet18}    &${256^2}$       &\textbf{88.43} &\textbf{89.35} &\underline{\textbf{78.28}} &\underline{\textbf{79.37}}       
                                                                 &\textbf{91.81} &\textbf{87.56} &\textbf{92.03} &\textbf{89.48}       
                                                                 &\underline{\textbf{99.25}} &\underline{\textbf{99.25}} &\underline{\textbf{99.25}} &\underline{\textbf{99.25}}            \\
\rowcolor[HTML]{EFEFEF}\textbf{HKG-ResNet34}    &${256^2}$       &\underline{\textbf{89.71}}  &\textbf{90.54} &\textbf{78.02} &\textbf{76.74}       
                                                                 &\textbf{92.44} &\textbf{88.55} &\textbf{92.21} &\textbf{90.16}       
                                                                 &\textbf{99.06} &\textbf{99.06} &\textbf{99.06} &\textbf{99.06}                        \\
\rowcolor[HTML]{EFEFEF}\textbf{HKG-VGG11}       &${256^2}$       &\textbf{85.71} &\textbf{88.21} &\textbf{75.78} &\textbf{76.19}       
                                                                 &\textbf{82.86} &\textbf{76.67} &\textbf{83.06} &\textbf{78.89}       
                                                                 &\textbf{98.63} &\textbf{98.63} &\textbf{98.63} &\textbf{98.63}                        \\  
\rowcolor[HTML]{EFEFEF}\textbf{HKG-DenseNet169} &${256^2}$       &\textbf{88.14} &\textbf{90.07} &\textbf{75.27} &\textbf{72.31}       
                                                                 &\underline{\textbf{92.69}} &\underline{\textbf{88.85}} &\underline{\textbf{92.48}} &\underline{\textbf{90.46} }      
                                                                 &\textbf{99.19} &\textbf{99.19} &\textbf{99.19} &\textbf{99.19}                        \\  
\midrule
MobNetv2 \cite{sandler2018mobilenetv2}           &${256^2}$       &84.00  &62.15  &69.98  &65.82    &88.30  &82.99  &88.08  &85.04       &94.13  &94.13  &94.13  &94.13                        \\
MobNetv3-S \cite{howard2019searching}            &${256^2}$       &76.71  &55.36  &63.38  &58.82    &85.38  &79.45  &85.56  &81.72       &92.38  &92.44  &92.38  &92.37                        \\
MobNetv3-L \cite{howard2019searching}            &${256^2}$       &77.43  &59.02  &61.39  &58.10    &86.87  &81.14  &86.43  &83.24       &93.44  &93.49  &93.44  &93.44                        \\
ShuNetv2-0.5 \cite{ma2018shufflenet}             &${256^2}$       &74.14  &58.67  &57.20  &53.04    &83.27  &77.15  &83.39  &79.36       &90.06  &90.06  &90.06  &90.06                        \\
ShuNetv2-1.0 \cite{ma2018shufflenet}             &${256^2}$       &74.86  &63.13  &57.98  &55.51    &83.00  &76.85  &83.06  &79.04       &90.56  &90.60  &90.56  &90.57                        \\
ShuNetv2-1.5 \cite{ma2018shufflenet}             &${256^2}$       &84.86  &63.12  &71.55  &67.07    &88.89  &83.68  &89.06  &85.84       &90.94  &90.94  &90.94  &90.92                        \\
ShuNetv2-2.0 \cite{ma2018shufflenet}             &${256^2}$       &73.29  &64.63  &58.35  &55.91    &82.53  &76.11  &81.94  &78.17       &89.94  &89.95  &89.94  &89.94                        \\
\rowcolor[HTML]{EFEFEF}\textbf{HKG-MobNetv2}     &${256^2}$       &\textbf{85.29} &\textbf{87.87} &\textbf{75.41} &\textbf{75.57}      
                                                                  &\textbf{89.70} &\textbf{85.00} &\textbf{89.73} &\textbf{86.98}       
                                                                  &\underline{\textbf{96.25}} &\underline{\textbf{96.25}} &\underline{\textbf{96.25}} &\underline{\textbf{96.25}}                        \\
\rowcolor[HTML]{EFEFEF}\textbf{HKG-ShuNetv2-1.5} &${256^2}$       &\underline{\textbf{86.57}} &\underline{\textbf{89.79}} &\underline{\textbf{73.55}} &\underline{\textbf{70.69}}       
                                                                  &\underline{\textbf{89.99}} &\underline{\textbf{85.23}} &\underline{\textbf{89.97}} &\underline{\textbf{87.21}}       
                                                                  &\textbf{93.13} &\textbf{93.14} &\textbf{93.13} &\textbf{93.13}                        \\   
\midrule
ViT-T-16 \cite{dosovitskiy2020image}             &${224^2}$       &86.86  &89.32  &76.81  &77.77     &91.23  &86.85  &91.09  &88.69       &96.19  &96.19  &96.19  &96.19                        \\
ViT-S-16 \cite{dosovitskiy2020image}             &${224^2}$       &87.14  &89.69  &76.89  &77.53     &91.73  &87.84  &91.78  &89.56       &96.38  &96.39  &96.38  &96.38                        \\
ViT-S-32 \cite{dosovitskiy2020image}             &${224^2}$       &87.43  &90.08  &77.02  &77.79     &91.93  &87.84  &92.08  &89.68       &96.50  &96.50  &96.50  &96.50                        \\
ViT-B-16 \cite{dosovitskiy2020image}             &${224^2}$       &87.57  &90.45  &74.90  &72.08     &92.05  &87.86  &92.06  &89.68       &96.81  &96.81  &96.81  &96.81                        \\
ViT-B-32 \cite{dosovitskiy2020image}             &${224^2}$       &87.86  &90.81  &75.17  &72.39     &92.65  &88.77  &92.46  &90.40       &97.25  &97.25  &97.25  &97.25                        \\
Swin-T-4-7 \cite{liu2021swin}                    &${224^2}$       &87.71  &90.63  &75.03  &72.24     &91.64  &87.35  &91.59  &89.16       &97.38  &97.40  &97.38  &97.38                        \\
Swin-S-4-7 \cite{liu2021swin}                    &${224^2}$       &87.86  &90.81  &75.17  &72.39     &91.90  &87.72  &91.43  &89.35       &97.63  &97.63  &97.63  &97.63                        \\
Swin-B-4-7 \cite{liu2021swin}                    &${224^2}$       &88.29  &91.11  &75.94  &73.69     &92.60  &88.58  &92.39  &90.25       &97.88  &97.88  &97.88  &97.87                        \\
Swin-B-4-12 \cite{liu2021swin}                   &${384^2}$       &88.43  &91.39  &75.90  &73.41     &92.88  &88.91  &93.09  &90.73       &99.19  &99.19  &99.19  &99.19                        \\  
\rowcolor[HTML]{EFEFEF}\textbf{HKG-ViT-B-32}     &${224^2}$       &\textbf{88.75} &\textbf{91.20} &\textbf{76.74} &\textbf{75.23}      
                                                                  &\textbf{93.09} &\textbf{89.54} &\textbf{93.01} &\textbf{91.09}       
                                                                  &\textbf{98.44} &\textbf{98.44} &\textbf{98.44} &\textbf{98.44}                        \\
\rowcolor[HTML]{EFEFEF}\textbf{HKG-Swin-B-4-12}  &${384^2}$       &\underline{\textbf{89.57}} &\underline{\textbf{92.35}} &\underline{\textbf{77.88}} &\underline{\textbf{76.66}}     
                                                                  &\underline{\textbf{93.18}} &\underline{\textbf{89.58}} &\underline{\textbf{93.40}} &\underline{\textbf{91.27}}       
                                                                  &\underline{\textbf{99.63}} &\underline{\textbf{99.63}} &\underline{\textbf{99.63}} &\underline{\textbf{99.62}}         \\   
\bottomrule
\end{tabular}}
\end{table}

In addition, we also report the accuracy of each cavitation state on three real-world cavitation datasets, see Table \ref{tab: Different Cavitation Fine Results}. It can be found that HKG can help different backbone networks to improve the accuracy of each cavitation state. It is worth noting that HKG can significantly improve the diagnosis results of incipient cavitation states without basically degrading the performance of other cavitation states. For example, the HKG assists ViT-B (32) to increase the incipient cavitation accuracy by \textbf{6.25}$\%$, \textbf{0.6}$\%$ and \textbf{1.5}$\%$ on three cavitation datasets, respectively.
\begin{table}[htbp]
\caption{Accuracy of various fine states on three real-world cavitation datasets, encompassing non-cavitation (non), choked flow cavitation (cho), constant cavitation (con), and incipient cavitation (inc). We show different representation learning methods including CNNs, LNNs and Transformers.}
\label{tab: Different Cavitation Fine Results}
\scriptsize
\setlength{\tabcolsep}{0.3mm}{
\begin{tabular}{l|cccc|cccc|cccc}
\toprule
\multicolumn{1}{c}{Dataset} & \multicolumn{4}{c}{Cavitation-Short}  & \multicolumn{4}{c}{Cavitation-Long} & \multicolumn{4}{c}{Cavitation-Noise}    \\ 
\midrule
\multicolumn{1}{c|}{Methods}& \multicolumn{1}{c}{cho} & \multicolumn{1}{c}{con}   & \multicolumn{1}{c}{inc}  & \multicolumn{1}{c|}{non} 
                            & \multicolumn{1}{c}{cho} & \multicolumn{1}{c}{con}   & \multicolumn{1}{c}{inc}  & \multicolumn{1}{c|}{non} 
                            & \multicolumn{1}{c}{cho} & \multicolumn{1}{c}{con}   & \multicolumn{1}{c}{inc}  & \multicolumn{1}{c}{non} \\ 
\midrule
ResNet34 \cite{he2016deep}                      &100.0   &96.11   &8.75    &100.0        &92.21   &92.24 	 &\underline{92.17}   &90.84         &98.50   &98.75 	&99.00 	  &99.00                 \\
VGG11 \cite{simonyan2014very}                   &85.71 	 &95.00   &0.00    &99.67        &81.04   &80.33 	 &81.93   &81.14                     &97.75   &98.00 	&97.25 	  &98.25                 \\
DenseNet169 \cite{huang2017densely}             &92.14 	 &100.0   &0.00    &100.0        &92.45   &91.96 	 &\underline{92.17}   &91.99         &98.50   &98.25 	&98.50 	  &\underline{99.50}                 \\
\rowcolor[HTML]{EFEFEF}\textbf{HKG-ResNet34}    &\underline{\textbf{100.0}} &\textbf{98.33} &\textbf{13.75} &\underline{\textbf{100.0}}       
                                                &\textbf{91.81}    &\underline{\textbf{92.82}}    &\textbf{91.87} 	&\textbf{92.35}        
                                                &\underline{\textbf{99.50}}    &\textbf{98.50} 	&\textbf{99.50} 	&\textbf{98.75}                 \\
\rowcolor[HTML]{EFEFEF}\textbf{HKG-VGG11}       &\textbf{93.57} &\textbf{87.22} &\underline{\textbf{25.00}} &\textbf{97.33}         
                                                &\textbf{83.25} &\textbf{82.78} &\textbf{83.73} 	&\textbf{82.47}         
                                                &\textbf{98.25} &\textbf{98.50} &\textbf{99.00} 	&\textbf{98.75}                 \\
\rowcolor[HTML]{EFEFEF}\textbf{HKG-DenseNet169} &\textbf{93.57} &\underline{\textbf{100.0}} &\textbf{7.50}  &\underline{\textbf{100.0}}        
                                                &\underline{\textbf{92.81}} &\textbf{92.71} 	&\textbf{91.47} 	&\underline{\textbf{92.92}}         
                                                &\textbf{98.75} 	&\underline{\textbf{99.50}} 	&\underline{\textbf{99.75}} 	&\textbf{98.75}                 \\
\midrule
MobileNetv2 \cite{sandler2018mobilenetv2}       &88.57 	 &91.67   &0.00    &99.67      &88.07 	&88.41 	&87.25 	&88.58                 &93.75 	&94.25 	  &93.25 	&\underline{95.25}                 \\
ShuNetv2-1.5 \cite{ma2018shufflenet}            &\underline{97.86} 	 &90.00   &0.00    &98.33   &89.40 	&88.64 	&89.36 	&88.86         &88.50 	&87.75 	  &94.25 	&93.25                \\
\rowcolor[HTML]{EFEFEF}\textbf{HKG-MobileNetv2} &\textbf{93.57}   &\textbf{88.33}  &\underline{\textbf{23.75}}  &\textbf{96.00}        
                                                &\underline{\textbf{90.36}}   &\textbf{89.64}  &\textbf{89.56}   &\textbf{89.34}         
                                                &\underline{\textbf{97.50}}   &\underline{\textbf{96.50}} 	&\underline{\textbf{95.75}} 	&\textbf{95.25}                 \\
\rowcolor[HTML]{EFEFEF}\textbf{HKG-ShuNetv2-1.5}&\textbf{96.43}   &\underline{\textbf{92.78}}   &\textbf{5.00}  &\underline{\textbf{100.0}} 
                                                &\textbf{90.00}   &\underline{\textbf{89.95}}   &\underline{\textbf{89.86}}   &\underline{\textbf{90.09}}        
                                                &\textbf{93.75}   &\textbf{92.75} 	&\textbf{92.75} 	&\textbf{93.25}                 \\
\midrule
ViT-B-32 \cite{dosovitskiy2020image}            &100.0 	 &94.44   &6.25    &100.0        &91.77   &92.86 	 &92.17   &93.04         &97.75   &98.25 	&96.50 	  &96.50                 \\
Swin-B-4-12 \cite{liu2021swin}                  &99.29 	 &96.11   &8.75    &100.0        &93.29   &92.54 	 &93.47   &93.04         &99.50   &98.75 	&98.75 	  &99.75                 \\
\rowcolor[HTML]{EFEFEF}\textbf{HKG-ViT-B-32}    &\underline{\textbf{100.0}}   &\textbf{94.44}   &\textbf{12.50}   &\underline{\textbf{100.0}}
                                                &\underline{\textbf{93.37}}   &\underline{\textbf{93.23}}   &\textbf{92.77}   &\textbf{92.68}         
                                                &\textbf{98.25} 	&\textbf{98.75} 	&\textbf{98.00} 	&\textbf{98.75}                 \\
\rowcolor[HTML]{EFEFEF}\textbf{HKG-Swin-B-4-12} &\underline{\textbf{100.0}}   &\underline{\textbf{97.78}}   &\underline{\textbf{13.75}}   &\underline{\textbf{100.0}}       
                                                &\textbf{93.29}   &\textbf{93.03}   &\underline{\textbf{94.18}}   &\underline{\textbf{93.10}}         
                                                &\underline{\textbf{100.0}} 	&\underline{\textbf{100.0}} 	&\underline{\textbf{99.50}} 	&\underline{\textbf{99.00}}                \\
\bottomrule
\end{tabular}}
\end{table}

Furthermore, we also show the results of different comparison methods on three real-world cavitation datasets, see Table \ref{tab: CavitationDatasets-FlowStatus}. It is clear from the table that HKG+ResNet34, HKG+ResNet18, HKG+DeneseNet169 and HKG+Swin-B-4-12 outperform the other methods in several critical metrics, especially in accuracy and F1 score, which show significant advantages. This suggests that the HKG method provides significant advantages for various backbones in cavitation intensity diagnosis, indicating the effective guidance of hierarchical knowledge on deep learning features.
\begin{table}[htbp]
\caption{Results of different evaluation metrics on three real-world cavitation datasets. Non-hierarchical and hierarchical methods from cavitation intensity recognition or fault diagnosis are used as comparative methods. The sliding window with a window size of 466944 and a step size of 466944.}
\label{tab: Compared Methods Cavitation Results}
\scriptsize
\setlength{\tabcolsep}{0.1mm}{
\begin{tabular}{l|c|cccc|cccc|cccc}
\toprule
\multicolumn{2}{c}{Dataset}      & \multicolumn{4}{c}{Cavitation-Short}  & \multicolumn{4}{c}{Cavitation-Long}  & \multicolumn{4}{c}{Cavitation-Noise}    \\ 
\midrule
\multicolumn{1}{c|}{Methods}    & \multicolumn{1}{c|}{\begin{tabular}[c]{@{}c@{}}Image \\ size\end{tabular}} & \multicolumn{1}{c}{Acc} & \multicolumn{1}{c}{Pre} & \multicolumn{1}{c}{Rec} & \multicolumn{1}{c|}{F1} 
                                & \multicolumn{1}{c}{Acc} & \multicolumn{1}{c}{Pre} & \multicolumn{1}{c}{Rec} & \multicolumn{1}{c|}{F1} 
                                & \multicolumn{1}{c}{Acc} & \multicolumn{1}{c}{Pre} & \multicolumn{1}{c}{Rec} & \multicolumn{1}{c}{F1} \\ 
\midrule
LiftingNet \cite{pan2017liftingnet}               &${256^2}$       & 85.29 & 82.75 & 75.89 & 73.95     & 88.43 & 86.31 & 87.97 & 87.05      & 95.86 & 94.62 & 95.84 & 95.20              \\
MIPLCNet \cite{pan2019novel}                      &${256^2}$       & 86.57 & 83.38 & 77.71 & 75.00     & 89.14 & 87.06 & 88.18 & 87.55      & 96.57 & 95.78 & 96.61 & 96.18                        \\
ResNet-APReLU \cite{zhao2020deep}                 &${256^2}$       & 86.86 & 84.04 & 77.89 & 75.56     & 90.71 & 88.83 & 90.91 & 89.70      & 97.29 & 96.77 & 97.49 & 97.12                         \\
LSTM-RDRN \cite{mohammad2023one}                  &${256^2}$       & 87.71 & 85.42 & \underline{78.50} & 76.72     & 91.14 & 89.36 & 91.55 & 90.32      & 98.71 & 98.40 & 98.53 & 98.46                         \\
BCNN  \cite{zhu2017b}                             &${256^2}$       & 81.71 & 78.63 & 72.25 & 70.03     & 85.71 & 82.42 & 85.09 & 83.56      & 92.14 & 89.75 & 92.71 & 91.03                         \\
\rowcolor[HTML]{EFEFEF}\textbf {HKG-ResNet18}     &${256^2}$       &\textbf{88.43} &\textbf{89.35} &\textbf{78.28} &\underline{\textbf{79.37}}       
                                                                   &\textbf{91.81} &\textbf{87.56} &\textbf{92.03} &\textbf{89.48}       
                                                                   &\textbf{99.25} &\textbf{99.25} &\textbf{99.25} &\textbf{99.25}  \\
                                                                   
\rowcolor[HTML]{EFEFEF}\textbf {HKG-ResNet34}     &${256^2}$       &\underline{\textbf{89.71}} &\textbf{90.54} &\textbf{78.02} &\textbf{76.74}       
                                                                   &\textbf{92.44} &\textbf{88.55} &\textbf{92.21} &\textbf{90.16}       
                                                                   &\textbf{99.06} &\textbf{99.06} &\textbf{99.06} &\textbf{99.06}  \\
                                                                   
\rowcolor[HTML]{EFEFEF}\textbf {HKG-DeneseNet169} &${256^2}$       &\textbf{88.14} &\textbf{90.07} &\textbf{75.27} &\textbf{72.31}       
                                                                   &\textbf{92.69} &\textbf{88.85} &\textbf{92.48} &\textbf{90.46}       
                                                                   &\textbf{99.19} &\textbf{99.19} &\textbf{99.19} &\textbf{99.19}   \\
                                                                   
\rowcolor[HTML]{EFEFEF}\textbf {HKG-Swin-B-4-12}  &${384^2}$       &\textbf{89.57} &\underline{\textbf{92.35}} &\textbf{77.88} &\textbf{76.66}     
                                                                   &\underline{\textbf{93.18}} &\underline{\textbf{89.58}} &\underline{\textbf{93.40}} &\underline{\textbf{91.27}}      
                                                                   &\underline{\textbf{99.63}} &\underline{\textbf{99.63}} &\underline{\textbf{99.63}} &\underline{\textbf{99.62}}   \\
\bottomrule
\end{tabular}}
\end{table}

\noindent\textbf{Results on PUB.} Table \ref{tab: PUB Results} shows the experimental results of different evaluation metrics. In general, HKG can help improve the diagnostic performance of various backbone networks. In particular, the HKG+ViT-S achieves the best performance of \textbf{98.92}$\%$ outperforming all SOAT methods and increasing average accuracy by \textbf{2.83}$\%$. And HKG also reaches the best precision, recall and F1-score of \textbf{98.3}$\%$, \textbf{99}$\%$ and \textbf{98.63}$\%$, respectively.

In detail, (1) \noindent\textbf{Health:} The diagnostic accuracy of HKG+ViT-S for bearing health is \textbf{100}$\%$, with an average improvement of \textbf{3.69}$\%$ compared to all SOAT methods. (2) \noindent\textbf{IR-1:} Although HKG+ViT-S, ResNet-APReLU, TDMSAE and SACL all obtain an accuracy of \textbf{98.21}$\%$ for IR-1, the performance of HKG is more superior for other states with ResNet18, ResNet34 and ViT-S. (3) \noindent\textbf{IR-2:} The results of HKG based on ResNet18/34 and ViT are all over \textbf{97}$\%$, which is better than all compared methods. In particular, HKG+ResNet34 achieves \textbf{100}$\%$ accuracy, which is on average \textbf{5.11}$\%$ better than comparison methods. (4) \noindent\textbf{IR-3:} HKG+ResNet18 and HKG+ViT-S all perform \textbf{100}$\%$, which an average improvement of \textbf{2.27}$\%$ over all baselines. (5) \noindent\textbf{OR-1:} HKG+ResNet34 and HKG+ViT all attain a performance of \textbf{99.11}$\%$. (6) \noindent\textbf{OR-2:} HKG+ResNet18, HKG+ResNet34 and HKG+ViT-S obtain \textbf{97.47}$\%$, \textbf{100}$\%$ and \textbf{98.73}$\%$ diagnostic results, which are \textbf{1.38}$\%$, \textbf{3.91}$\%$ and \textbf{2.65}$\%$ improvement compared to SOAT methods, respectively.
\begin{table}[htbp]
\caption{Different evaluation metrics results on PUB dataset.}
\label{tab: PUB Results}
\scriptsize
\setlength{\tabcolsep}{0.1mm}{
\begin{tabular}{l|cc|cccccc|cccc}
\toprule
\multicolumn{1}{c}{\multirow{2}{*}{Methods}} & \multirow{2}{*}{Backbones} 
                                              & \multicolumn{1}{c}{\multirow{2}{*}{\begin{tabular}[c]{@{}c@{}}Image\\ size\end{tabular}}} 
                                              & \multicolumn{6}{c}{Accuracy of fine states}                                                                                                                        
                                              & \multicolumn{4}{c}{Overall}\\ 
\cmidrule{4-13} 
\multicolumn{1}{c}{}    &  & \multicolumn{1}{c}{} 
                        & \multicolumn{1}{c}{Health}
                        & \multicolumn{1}{c}{IR-1} 
                        & \multicolumn{1}{c}{IR-2} 
                        & \multicolumn{1}{c}{IR-3} 
                        & \multicolumn{1}{c}{OR-1} 
                        & \multicolumn{1}{c}{OR-2}
                        & Acc   & Pre   & Rec   & F1  \\ 
\midrule
DATMMD \cite{li2020intelligent}     & \multicolumn{1}{c|}{LeNet}          &${256^2}$        &90.63  &93.75  &93.75  &100.0  &93.75  &94.94  &93.52  &92.46  &94.47  &93.38      \\
LiftingNet \cite{pan2017liftingnet} & \multicolumn{1}{c|}{AlexNet}        &${256^2}$        &95.83 	&90.18 	&93.75 	&93.75 	&94.64 	&92.41  &93.30  &91.18  &93.43  &92.09      \\
MIP-LCNet \cite{pan2019novel}       & \multicolumn{1}{c|}{AlexNet}        &${256^2}$        &97.92 	&89.29 	&95.83 	&93.75 	&94.64 	&96.20  &93.48  &91.68  &94.61  &92.96      \\
ResNet18 \cite{he2016deep}          & \multicolumn{1}{c|}{ResNet18}       &${256^2}$        &94.79 	&96.43 	&93.75 	&100.0 	&95.54 	&94.94  &95.46  &94.86  &95.91  &95.36      \\
ResNet-APReLU \cite{zhao2020deep}   & \multicolumn{1}{c|}{ResNet18}       &${256^2}$        &97.92 	&98.21 	&91.67 	&100.0 	&96.43 	&96.20  &96.76  &95.07  &96.74  &95.78      \\
ADMTL \cite{li2023attention}        & \multicolumn{1}{c|}{ResNet18}       &${256^2}$        &95.83 	&95.54 	&97.92 	&93.75 	&98.21 	&96.20  &96.54  &95.13  &96.24  &95.66      \\
RDRN \cite{mohammad2023one}         & \multicolumn{1}{c|}{ResNet18}       &${256^2}$        &97.92 	&96.43 	&95.83 	&100.0 	&98.21 	&94.94  &96.98  &94.32  &97.22  &95.57      \\
TDMSAE \cite{yu2023tdmsae}          & \multicolumn{1}{c|}{ResNet50}       &${256^2}$        &93.75 	&98.21 	&95.83 	&93.75 	&99.11 	&98.73  &97.19  &95.50  &96.56  &95.96      \\
LRSADTLM \cite{yu2023adaptive}      & \multicolumn{1}{c|}{Transformer}    &${224^2}$        &97.92 	&97.32 	&95.83 	&100.0 	&99.11 	&96.20  &97.62  &96.40  &97.73  &97.00      \\
SACL \cite{cui2024self}             & \multicolumn{1}{c|}{Transformer}    &${224^2}$        &97.92 	&98.21 	&93.75 	&100.0 	&97.32 	&97.47  &97.41  &95.83  &97.45  &96.51      \\
TS-TCC \cite{eldele2023self}        & \multicolumn{1}{c|}{Transformer}    &${224^2}$        &98.96 	&97.32 	&95.83 	&100.0  &97.32 	&98.73  &97.84  &96.56  &98.03  &97.23      \\
\midrule
\rowcolor[HTML]{EFEFEF}
\multicolumn{1}{c|}{}  
& \multicolumn{1}{c|}{\textbf{ResNet18}}            &${256^2}$  &\textbf{96.88} 
                                                                &\textbf{94.64} 
                                                                &\textbf{97.92} 
                                                                &\underline{\textbf{100.0}} 
                                                                &\textbf{97.32} 
                                                                &\textbf{97.47}        
                                                                &\textbf{96.76}       
                                                                &\textbf{94.59}       
                                                                &\textbf{97.37}       
                                                                &\textbf{95.79}      \\
\rowcolor[HTML]{EFEFEF}
\multicolumn{1}{c|}{\multirow{-1}{*}{\textbf{HKG}}} 
& \multicolumn{1}{c|}{\textbf{ResNet34}}            &${256^2}$ &\textbf{95.83} 
                                                               &\textbf{97.32} 
                                                               &\underline{\textbf{100.0}} 
                                                               &\textbf{93.75} 
                                                               &\underline{\textbf{99.11}} 
                                                               &\underline{\textbf{100.0}}  
                                                               &\textbf{98.06}       
                                                               &\textbf{97.40}       
                                                               &\textbf{97.67}
                                                               &\textbf{97.53}      \\
\rowcolor[HTML]{EFEFEF}
\multicolumn{1}{c|}{}    
& \multicolumn{1}{c|}{\textbf{ViT-S (32)}}         &${224^2}$ &\underline{\textbf{100.0}} 
                                                              &\underline{\textbf{98.21}} 
                                                              &\textbf{97.92} 
                                                              &\underline{\textbf{100.0}} 
                                                              &\underline{\textbf{99.11}} 
                                                              &\textbf{98.73}    
                                                              &\underline{\textbf{98.92}}       
                                                              &\underline{\textbf{98.30}}       
                                                              &\underline{\textbf{99.00}}       
                                                              &\underline{\textbf{98.63}}      \\
\bottomrule
\end{tabular}}
\end{table}

\begin{figure*}[htbp]
\centering
% \subfigure[Different word embeddings]{
% \begin{minipage}[t]{0.2\linewidth}
% \centering
% \includegraphics[width=0.9\textwidth,height=22mm]{fig/Experiments/DifferentWordEmbedding.pdf}
% %\caption{fig1}
% \end{minipage}%
% }%
\subfigure[Parameter sensitivity]{
\begin{minipage}[t]{0.25\linewidth}
\centering
\includegraphics[width=0.9\textwidth,height=28mm]{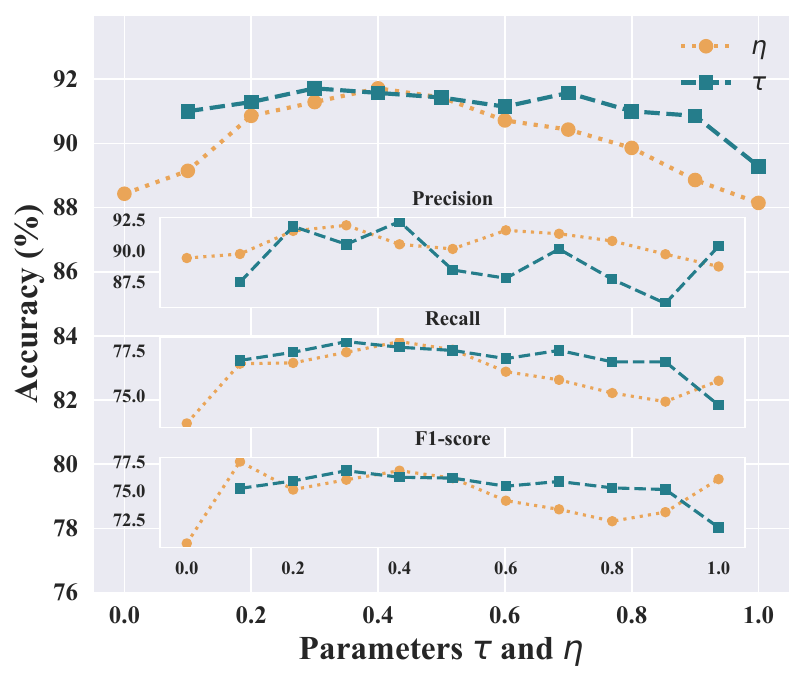}
%\caption{fig2}
\end{minipage}%
}%
\subfigure[STFT parameter analysis]{
\begin{minipage}[t]{0.25\linewidth}
\centering
\includegraphics[width=0.9\textwidth,height=28mm]{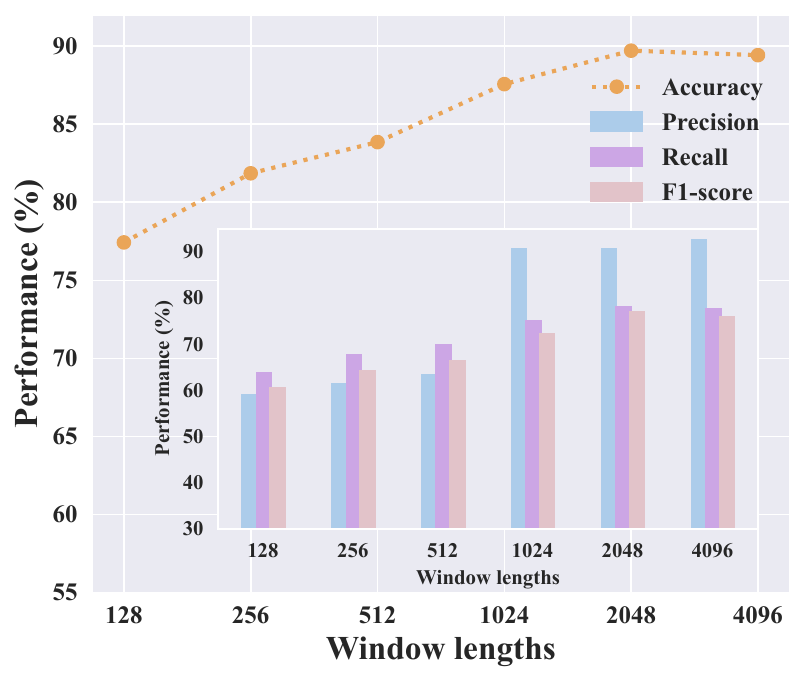}
%\caption{fig1}
\end{minipage}%
}%
\subfigure[Window size analysis]{
\begin{minipage}[t]{0.25\linewidth}
\centering
\includegraphics[width=0.9\textwidth,height=28mm]{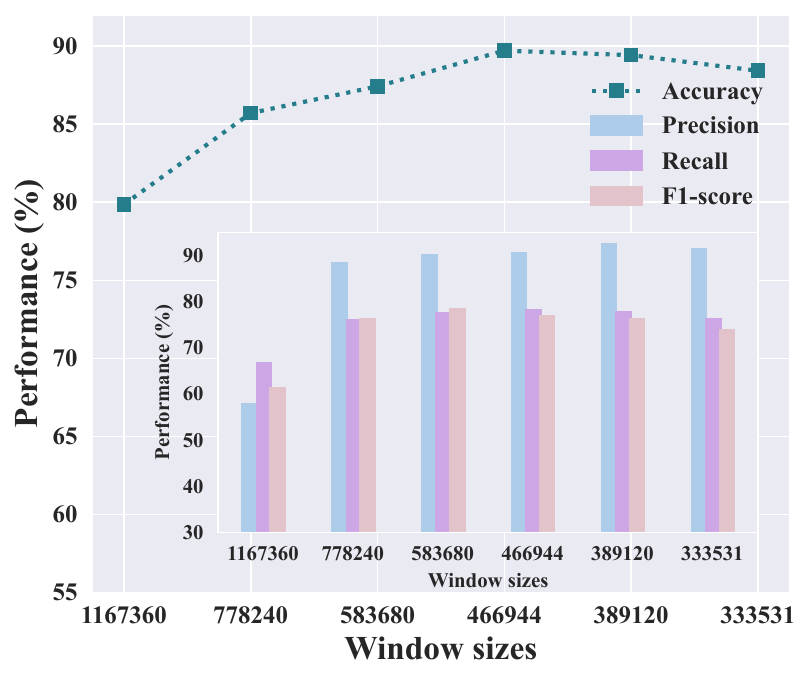}
%\caption{fig1}
\end{minipage}%
}%
\subfigure[Downsampling effects]{
\begin{minipage}[t]{0.25\linewidth}
\centering
\includegraphics[width=0.9\textwidth,height=28mm]{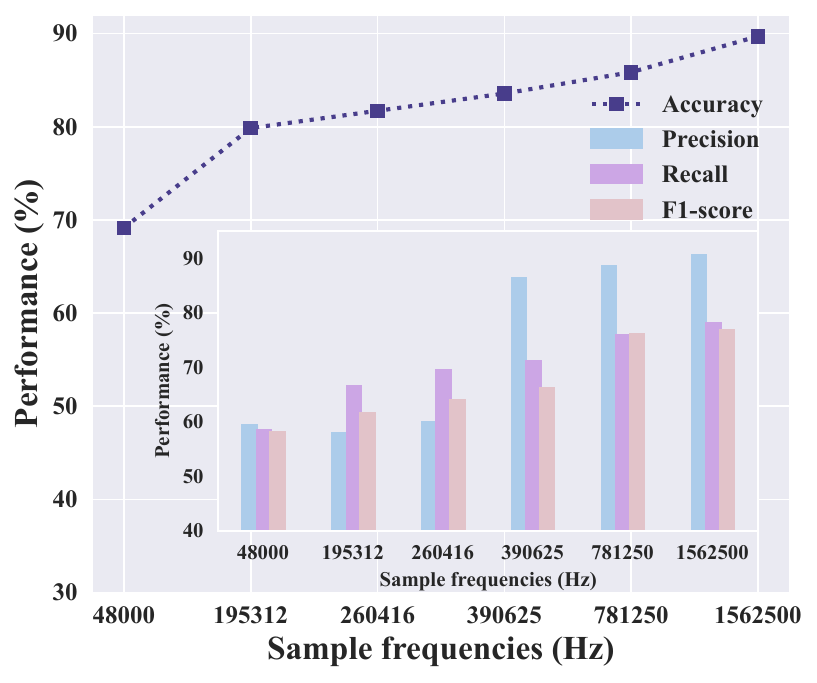}
%\caption{fig1}
\end{minipage}%
}%
\centering
\caption{Different evaluation metrics results of various ablation experiments on Cavitation-Short. (a)-(b) and (d) are all performed with a window size of 466944 and a step size of 466944.} 
\label{fig: ablation analysis}
\end{figure*}

\noindent\textbf{Takeaways:} The proposed HKG achieves significant improvements in results for various backbone networks on four real-world datasets. This is motivated by two following reasons. Firstly, the hierarchical knowledge learning module of HKG explicitly learns the hierarchical relationships between target classes. Secondly, the global hierarchical classifier effectively constrains and guides the learned deep features from representation learning.

\subsection{Ablation Analysis}
\label{subsec: Ablation Analysis}
\noindent\textbf{Effects of correlation matrices.} We report the results with different correlation matrices for our method in Table \ref{tab: Correlation Matrix}. It can be found that the Re-HEKCM for HKG significantly outperforms the other correlation matrices (SCM, HEKCM and B-HEKCM), which shows the Re-HEKCM effectively overcomes the issues mentioned in Section \ref{subsec: Hierarchical Knowledge Learning}. Meanwhile, the performance of HKG with Re-HEKCM, B-HEKCM and HEKCM are superior to HKG with SCM. It demonstrates that correlation matrices (i.e., Re-HEKCM, B-HEKCM and HEKCM) obtained by embedding hierarchical knowledge between classes in SCM are more conducive to guiding deep features.
\begin{table}[htbp]
\caption{Effects of correlation matrices on Cavitation-Short.}
\label{tab: Correlation Matrix}
\scriptsize
\setlength{\tabcolsep}{0.9mm}{
\begin{tabular}{cccc|cccc}
\toprule
\multicolumn{4}{c}{HKG + ResNet-34}  & \multicolumn{4}{c}{Evaluation Metrics} \\ 
\midrule
 SCM         & HEKCM       & B-HEKCM     & Re-HEKCM          & Accuracy   & Precision   & Recall   & F1-score  \\ 
\midrule
\ding{55}    &\ding{55}    &\ding{55}    &\ding{55}          &88.57\textsubscript{\raisebox{0.2ex}{\textcolor[RGB]{255,255,255}{\scalebox{0.5}{$\uparrow$}}}\textcolor[RGB]{255,255,255}{\scalebox{0.65}{0.00}}}    
                                                             &91.47\textsubscript{\raisebox{0.2ex}{\textcolor[RGB]{255,255,255}{\scalebox{0.5}{$\uparrow$}}}\textcolor[RGB]{255,255,255}{\scalebox{0.65}{0.00}}}        
                                                             &76.22\textsubscript{\raisebox{0.2ex}{\textcolor[RGB]{255,255,255}{\scalebox{0.5}{$\uparrow$}}}\textcolor[RGB]{255,255,255}{\scalebox{0.65}{0.00}}}     
                                                             &74.00\textsubscript{\raisebox{0.2ex}{\textcolor[RGB]{255,255,255}{\scalebox{0.5}{$\uparrow$}}}\textcolor[RGB]{255,255,255}{\scalebox{0.65}{0.00}}}     \\
\ding{51}    &\ding{55}    &\ding{55}    &\ding{55}          &89.00\textsubscript{\raisebox{0.2ex}{\textcolor[RGB]{33,170,229}{\scalebox{0.5}{$\uparrow$}}}\textcolor[RGB]{33,170,229}{\scalebox{0.65}{0.43}}}     
                                                             &91.80\textsubscript{\raisebox{0.2ex}{\textcolor[RGB]{33,170,229}{\scalebox{0.5}{$\uparrow$}}}\textcolor[RGB]{33,170,229}{\scalebox{0.65}{0.33}}}             
                                                             &76.98\textsubscript{\raisebox{0.2ex}{\textcolor[RGB]{33,170,229}{\scalebox{0.5}{$\uparrow$}}}\textcolor[RGB]{33,170,229}{\scalebox{0.65}{0.76}}}     
                                                             &75.27\textsubscript{\raisebox{0.2ex}{\textcolor[RGB]{33,170,229}{\scalebox{0.5}{$\uparrow$}}}\textcolor[RGB]{33,170,229}{\scalebox{0.65}{1.27}}}      \\
\ding{55}    &\ding{51}    &\ding{55}    &\ding{55}          &89.29\textsubscript{\raisebox{0.2ex}{\textcolor[RGB]{33,170,229}{\scalebox{0.5}{$\uparrow$}}}\textcolor[RGB]{33,170,229}{\scalebox{0.65}{0.72}}}       
                                                             &92.00\textsubscript{\raisebox{0.2ex}{\textcolor[RGB]{33,170,229}{\scalebox{0.5}{$\uparrow$}}}\textcolor[RGB]{33,170,229}{\scalebox{0.65}{0.53}}}         
                                                             &77.43\textsubscript{\raisebox{0.2ex}{\textcolor[RGB]{33,170,229}{\scalebox{0.5}{$\uparrow$}}}\textcolor[RGB]{33,170,229}{\scalebox{0.65}{1.21}}}     
                                                             &75.89\textsubscript{\raisebox{0.2ex}{\textcolor[RGB]{33,170,229}{\scalebox{0.5}{$\uparrow$}}}\textcolor[RGB]{33,170,229}{\scalebox{0.65}{1.89}}}      \\
\ding{55}    &\ding{55}    &\ding{51}    &\ding{55}          &89.57\textsubscript{\raisebox{0.2ex}{\textcolor[RGB]{33,170,229}{\scalebox{0.5}{$\uparrow$}}}\textcolor[RGB]{33,170,229}{\scalebox{0.65}{1.00}}}       
                                                             &\textbf{92.25}\textsubscript{\raisebox{0.2ex}{\textcolor[RGB]{33,170,229}{\scalebox{0.5}{$\uparrow$}}}\textcolor[RGB]{33,170,229}{\scalebox{0.65}{0.78}}}         
                                                             &77.88\textsubscript{\raisebox{0.2ex}{\textcolor[RGB]{33,170,229}{\scalebox{0.5}{$\uparrow$}}}\textcolor[RGB]{33,170,229}{\scalebox{0.65}{1.66}}}     
                                                             &76.56\textsubscript{\raisebox{0.2ex}{\textcolor[RGB]{33,170,229}{\scalebox{0.5}{$\uparrow$}}}\textcolor[RGB]{33,170,229}{\scalebox{0.65}{2.56}}}     \\
\rowcolor[HTML]{EFEFEF}
\ding{55}    &\ding{55}    &\ding{55}    &\ding{51}          &\textbf{89.71}\textsubscript{\raisebox{0.2ex}{\textcolor[RGB]{33,170,229}{\scalebox{0.5}{$\uparrow$}}}\textcolor[RGB]{33,170,229}{\scalebox{0.65}{1.14}}}       
                                                             &90.54\textsubscript{\raisebox{0.2ex}{\textcolor[RGB]{33,170,229}{\scalebox{0.5}{$\downarrow$}}}\textcolor[RGB]{33,170,229}{\scalebox{0.65}{0.93}}}        
                                                             &\textbf{78.02}\textsubscript{\raisebox{0.2ex}{\textcolor[RGB]{33,170,229}{\scalebox{0.5}{$\uparrow$}}}\textcolor[RGB]{33,170,229}{\scalebox{0.65}{1.80}}}     
                                                             &\textbf{76.74}\textsubscript{\raisebox{0.2ex}{\textcolor[RGB]{33,170,229}{\scalebox{0.5}{$\uparrow$}}}\textcolor[RGB]{33,170,229}{\scalebox{0.65}{2.74}}}     \\
\bottomrule
\end{tabular}}
\end{table}

\noindent\textbf{Analysis of GCN layer number.} From Table \ref{tab: GCN Layers}, it can be found that as the number of graph convolutional layers increases, the performance of our model initially increases and then decreases. When the number of layers is 0, our model is ResNet34. The possible reason for the performance drop is the oversmoothing between nodes when the GCN is deep.
\begin{table}[htbp]
\caption{Effects of depth of GCN on Cavitation-Short.}
\label{tab: GCN Layers}
\scriptsize
\setlength{\tabcolsep}{3.5mm}{
\begin{tabular}{c|cccc}
\toprule
\multirow{2}{*}{GCN Layers} & \multicolumn{4}{c}{Evaluation Metrics}   \\ 
\cmidrule{2-5} 
                            & Accuracy & Precision & Recall & F1-score \\ 
\midrule
0-layer                     &88.57\textsubscript{\raisebox{0.2ex}{\textcolor[RGB]{255,255,255}{\scalebox{0.5}{$\uparrow$}}}\textcolor[RGB]{255,255,255}{\scalebox{0.65}{0.00}}}      
                            &91.47\textsubscript{\raisebox{0.2ex}{\textcolor[RGB]{255,255,255}{\scalebox{0.5}{$\uparrow$}}}\textcolor[RGB]{255,255,255}{\scalebox{0.65}{0.00}}}      
                            &76.22\textsubscript{\raisebox{0.2ex}{\textcolor[RGB]{255,255,255}{\scalebox{0.5}{$\uparrow$}}}\textcolor[RGB]{255,255,255}{\scalebox{0.65}{0.00}}}      
                            &74.00\textsubscript{\raisebox{0.2ex}{\textcolor[RGB]{255,255,255}{\scalebox{0.5}{$\uparrow$}}}\textcolor[RGB]{255,255,255}{\scalebox{0.65}{0.00}}}          \\
1-layer                     &89.00\textsubscript{\raisebox{0.2ex}{\textcolor[RGB]{33,170,229}{\scalebox{0.5}{$\uparrow$}}}\textcolor[RGB]{33,170,229}{\scalebox{0.65}{0.43}}}       
                            &91.80\textsubscript{\raisebox{0.2ex}{\textcolor[RGB]{33,170,229}{\scalebox{0.5}{$\uparrow$}}}\textcolor[RGB]{33,170,229}{\scalebox{0.65}{0.33}}}       
                            &76.98\textsubscript{\raisebox{0.2ex}{\textcolor[RGB]{33,170,229}{\scalebox{0.5}{$\uparrow$}}}\textcolor[RGB]{33,170,229}{\scalebox{0.65}{0.76}}}       
                            &75.27\textsubscript{\raisebox{0.2ex}{\textcolor[RGB]{33,170,229}{\scalebox{0.5}{$\uparrow$}}}\textcolor[RGB]{33,170,229}{\scalebox{0.65}{1.27}}}           \\
\rowcolor[HTML]{EFEFEF}
2-layer                     &\textbf{89.71}\textsubscript{\raisebox{0.2ex}{\textcolor[RGB]{33,170,229}{\scalebox{0.5}{$\uparrow$}}}\textcolor[RGB]{33,170,229}{\scalebox{0.65}{1.14}}}       
                            &90.54\textsubscript{\raisebox{0.2ex}{\textcolor[RGB]{33,170,229}{\scalebox{0.5}{$\downarrow$}}}\textcolor[RGB]{33,170,229}{\scalebox{0.65}{0.93}}}       
                            &\textbf{78.02}\textsubscript{\raisebox{0.2ex}{\textcolor[RGB]{33,170,229}{\scalebox{0.5}{$\uparrow$}}}\textcolor[RGB]{33,170,229}{\scalebox{0.65}{1.80}}}       
                            &\textbf{76.74}\textsubscript{\raisebox{0.2ex}{\textcolor[RGB]{33,170,229}{\scalebox{0.5}{$\uparrow$}}}\textcolor[RGB]{33,170,229}{\scalebox{0.65}{2.74}}}          \\
3-layer                     &89.43\textsubscript{\raisebox{0.2ex}{\textcolor[RGB]{33,170,229}{\scalebox{0.5}{$\uparrow$}}}\textcolor[RGB]{33,170,229}{\scalebox{0.65}{0.86}}}       
                            &\textbf{92.17}\textsubscript{\raisebox{0.2ex}{\textcolor[RGB]{33,170,229}{\scalebox{0.5}{$\uparrow$}}}\textcolor[RGB]{33,170,229}{\scalebox{0.65}{0.70}}}       
                            &77.57\textsubscript{\raisebox{0.2ex}{\textcolor[RGB]{33,170,229}{\scalebox{0.5}{$\uparrow$}}}\textcolor[RGB]{33,170,229}{\scalebox{0.65}{1.35}}}       
                            &76.02\textsubscript{\raisebox{0.2ex}{\textcolor[RGB]{33,170,229}{\scalebox{0.5}{$\uparrow$}}}\textcolor[RGB]{33,170,229}{\scalebox{0.65}{2.02}}}           \\ 
\bottomrule
\end{tabular}}
\end{table} 

\noindent\textbf{Impacts of word embeddings.} We compare the performance of our model using different word embedding methods on Cavitation-Short. It is clear from Table \ref{tab: Word Embedding} that the three different word embedding methods have a negligible effect on the performance of our model. It indicates that the performance of our methods does not depend entirely on the semantic information of the word embeddings. In addition, the reason that powerful word embeddings achieve higher performance is that learned word embeddings from publicly available semantic repositories keep topological structure between classes.
\begin{table}[htbp]
\caption{Results of word embedding on Cavitation-Short.}
\label{tab: Word Embedding}
\scriptsize
\setlength{\tabcolsep}{3.5mm}{
\begin{tabular}{l|cccc}
\toprule
\multirow{2}{*}{Types}      & \multicolumn{4}{c}{Evaluation Metrics}   \\ 
\cmidrule{2-5} 
                            & Accuracy & Precision & Recall & F1-score \\ 
\midrule
FastText \cite{joulin2016bag}&88.57\textsubscript{\raisebox{0.2ex}{\textcolor[RGB]{255,255,255}{\scalebox{0.5}{$\uparrow$}}}\textcolor[RGB]{255,255,255}{\scalebox{0.65}{0.00}}}      
                             &91.47\textsubscript{\raisebox{0.2ex}{\textcolor[RGB]{255,255,255}{\scalebox{0.5}{$\uparrow$}}}\textcolor[RGB]{255,255,255}{\scalebox{0.65}{0.00}}}      
                             &76.22\textsubscript{\raisebox{0.2ex}{\textcolor[RGB]{255,255,255}{\scalebox{0.5}{$\uparrow$}}}\textcolor[RGB]{255,255,255}{\scalebox{0.65}{0.00}}}      
                             &74.00\textsubscript{\raisebox{0.2ex}{\textcolor[RGB]{255,255,255}{\scalebox{0.5}{$\uparrow$}}}\textcolor[RGB]{255,255,255}{\scalebox{0.65}{0.00}}}\\
                            
GoogleNews \cite{mikolov2013efficient}&89.00\textsubscript{\raisebox{0.2ex}{\textcolor[RGB]{33,170,229}{\scalebox{0.5}{$\uparrow$}}}\textcolor[RGB]{33,170,229}{\scalebox{0.65}{0.43}}}       
                                      &91.80\textsubscript{\raisebox{0.2ex}{\textcolor[RGB]{33,170,229}{\scalebox{0.5}{$\uparrow$}}}\textcolor[RGB]{33,170,229}{\scalebox{0.65}{0.33}}}       
                                      &76.98\textsubscript{\raisebox{0.2ex}{\textcolor[RGB]{33,170,229}{\scalebox{0.5}{$\uparrow$}}}\textcolor[RGB]{33,170,229}{\scalebox{0.65}{0.76}}}       
                                      &75.27\textsubscript{\raisebox{0.2ex}{\textcolor[RGB]{33,170,229}{\scalebox{0.5}{$\uparrow$}}}\textcolor[RGB]{33,170,229}{\scalebox{0.65}{1.27}}}\\
\rowcolor[HTML]{EFEFEF}
GloVe \cite{pennington2014glove}&\textbf{89.43}\textsubscript{\raisebox{0.2ex}{\textcolor[RGB]{33,170,229}{\scalebox{0.5}{$\uparrow$}}}\textcolor[RGB]{33,170,229}{\scalebox{0.65}{0.86}}}       
                                &\textbf{92.17}\textsubscript{\raisebox{0.2ex}{\textcolor[RGB]{33,170,229}{\scalebox{0.5}{$\uparrow$}}}\textcolor[RGB]{33,170,229}{\scalebox{0.65}{0.70}}}       
                                &\textbf{77.57}\textsubscript{\raisebox{0.2ex}{\textcolor[RGB]{33,170,229}{\scalebox{0.5}{$\uparrow$}}}\textcolor[RGB]{33,170,229}{\scalebox{0.65}{1.35}}}       
                                &\textbf{76.02}\textsubscript{\raisebox{0.2ex}{\textcolor[RGB]{33,170,229}{\scalebox{0.5}{$\uparrow$}}}\textcolor[RGB]{33,170,229}{\scalebox{0.65}{2.02}}}\\ 
\bottomrule
\end{tabular}}
\end{table}

\noindent\textbf{Parameters ($\tau$ $\&$ $\eta$) sensitivity.} We analyze the impact of different values of threshold $\tau$ (Equation \ref{eq: B-HKCM}) and scaling factor $\eta$ (Equation \ref{eq: Re-HKCM}) on for the HKG on Cavitation-Short, see Figure \ref{fig: ablation analysis}a. For $\tau$, HKG accuracy is improved when a small amount of edge noise is filtered. When a large amount of edge noise is filtered causing neighbor information being lost, HKG accuracy decreases. Where, HKG does not converge for $\tau  = 0$. The optimal value of $\tau$ is 0.3. For $\eta$, HKG achieves the best accuracy with $\eta  = 0.4$. When $\eta$ is set too small, the node only considers its own features. Conversely, $\eta$ is set too large, the node considers excessive neighbour information, which occurs as an over-smoothing issue.

\noindent\textbf{STFT parameter analysis.} We consider the effect of the window length of the STFT for the performance of HKG. In all experiments, the step length is one-quarter of the window length. From Figure \ref{fig: ablation analysis}b, it can be seen that the window length has a very significant impact on the results and the most appropriate window length is 2048.

\noindent\textbf{Window size analysis.} We compare the results of our method with different window sizes on Cavitation-Short. It is clear from Figure \ref{fig: ablation analysis}c that the window size is sensitive to the performance of HKG and the best window size is 466944.

\noindent\textbf{Downsampling effects.} In practical applications, the ability to recognize signals obtained from low-level sensors is very important and challenging. Our method is evaluated under the original sample frequency ($Fs$ = \SI{1562500}{Hz}), one-half, one-quarter, one-sixth, one-eighth of the original sample frequency (\SI{781250}{Hz}, \SI{390625}{Hz}, \SI{260416}{Hz}, \SI{195312}{Hz}) and the maximum frequency a mobile phone can withstand (\SI{48000}{Hz} $\approx Fs/32$). As shown in Figure \ref{fig: ablation analysis}d, although the performance of HKG gradually reduces as the sampling frequency decreases, the accuracy always remains above \textbf{80}$\%$. Moreover, HKG also reaches \textbf{69}$\%$ accuracy with the maximum sample frequency of mobile phones.

\section{Related Work}
\label{sec: related work}
\noindent\textbf{Fault Intensity Diagnosis.} Fault intensity diagnosis (FID) is usually regarded as a specific recognition problem based on a fine-grained fault classification in terms of signals. Signal-based FID can be broadly organized into two classes: convolutional neural networks (CNNs) and transformers. For CNN methods, Zhou et al. in \cite{li2020intelligent} proposed an intelligent fault diagnosis method based on feature space distance extractor and feature domain mismatch extractor. Pan et al. in \cite{pan2017liftingnet} designed a novel CNN consisting of split layers inspired by the second generation wavelet transform. Pan et al. in \cite{pan2019novel} developed an intelligent fault detection method using multiscale inner product and local feature connection. Li et al. in \cite{li2023attention} presented a novel attention-based deep meta-transfer learning for fine-grained fault diagnosis. Arta et al. \cite{mohammad2023one} recommend a deep residual network modulated by a long- and short-term memory network for vehicle fault diagnosis. Yu et al. in \cite{yu2023tdmsae} presented a decoupled multi-scale autoencoder fault diagnosis model. For transformer methods, Yu et al. in \cite{yu2023adaptive} proposed a deep transfer method for fault diagnosis by combining time-frequency analysis, residual networks and self-attention. Cui et al. in \cite{cui2024self} introduced the transformer as the backbone of a contrastive learning fault diagnosis model. However, the above methods do not consider the hierarchical structure information between target classes.
 
\noindent\textbf{Hierarchical Classification.} Considering the hierarchical information of target classes is an active topic in machine learning application domains \cite{mousavi2017hierarchical}. According to the target class belonging to one or more paths of hierarchical structure, hierarchical classification can be viewed as a special kind of multi-label classification task. \cite{li2022deep}. For deep learning methods, existing hierarchical classifications can be divided into three families: label embedding, hierarchical losses and hierarchical structures. For label embedding methods \cite{chen2019multi, bengio2010label}, the hierarchical information across labels is mapped into hierarchical semantic vectors containing relative position information. For hierarchical losses \cite{li2022deep,bertinetto2020making}, which emphasizes the consistency between the prediction results and the class hierarchy that is often employed in multi-label classification tasks. Hierarchical structures \cite{wu2020learning,meng2019hats,ma2019hierarchical} aim to better adapt specifically designed deep neural network structures to the class hierarchy of a particular task. In this work, our proposed HKG is one of the label embedding methods.

\section{Conclusions}
Fault intensity diagnosis FID is essential for monitoring complex industrial systems. In this paper, we propose HKG, a novel end-to-end knowledge- and data-driven FID approach that uses graph convolutional neural networks to map hierarchical topological graphs of label representations together with learned deep features from representation learning to a set of interdependent global hierarchical classifiers. To explicitly model the hierarchical dependencies among classes, we design a novel re-weighted hierarchical knowledge correlation matrix scheme by embedding hierarchical knowledge in a data-driven statistical correlation matrix. The scheme can effectively mitigate over-fitting and over-smoothing issues by balancing the weights between nodes and their neighborhoods. The HKG outperforms the SOTA methods on four real-world datasets. The ablation experiments clearly show that HKG successfully meets the FID requirements of complex industrial systems in the real world, providing insights for extending the method to other industrial applications.

\section*{Acknowledgements}
\label{sec:acknowledgements}
This research is supported by Xidian-FIAS International Joint Research Center (Y. S.), by the AI grant at FIAS through SAMSON AG (J. F., K. Z.), by the CUHK-Shenzhen university development fund under grant No. UDF01003041 and the BMBF funded KISS consortium (05D23RI1) in the ErUM-Data action plan (K. Z.), by SAMSON AG (D. V., T. S., A. W.), by the Walter GreinerGesellschaft zur F\"orderung der physikalischen Grundla - genforschung e.V. through the Judah M. Eisenberg Lau-reatus Chair at Goethe Universit\"at Frankfurt am Main (H. S.), by the NVIDIA GPU grant through NVIDIA Corporation (K. Z.), by the National Natural Science Foundation of China under No. 62372358 and No. 62301395 (S. G.), and by the Shaanxi Province Postdoctoral Science Foundation under NO. 2023BSHEDZZ177 (S. G.).

\clearpage
\bibliographystyle{ACM-Reference-Format}
\bibliography{main}

\clearpage
\appendix
\setcounter{table}{0}
\setcounter{figure}{0}
\setcounter{equation}{0}
\renewcommand{\thetable}{A\arabic{table}}
\renewcommand{\thefigure}{A\arabic{figure}}
\renewcommand{\theequation}{A.\arabic{equation}}

\section{PRELIMINARIES}
\label{sec: Appendix Preliminaries}

\subsection{Definition of Cavitation Intensity}
\label{app: Cavitation Intensity Definition}
\begin{figure}[htbp]
    \centering
    \includegraphics[width=0.4\textwidth,height=40mm]{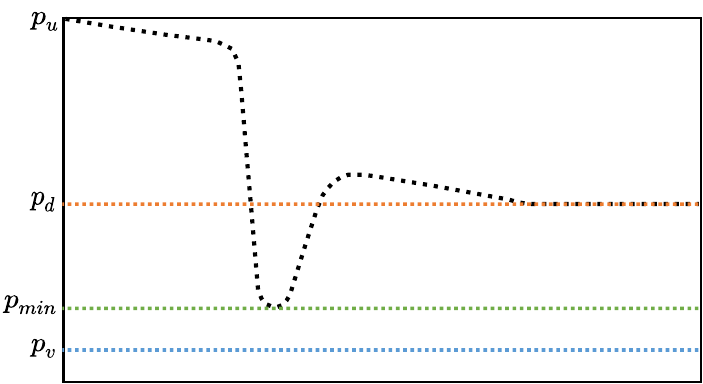}
    \caption{Pressure flow model in a valve. The solid black, dotted orange, dotted green and dotted blue lines are represented by the upstream pressure ${p}_{u}$, the downstream pressure ${p}_{d}$, the minimum pressure ${p}_{min}$ and the vapor pressure ${p}_{v}$, respectively.}
    \label{fig: civitation knowledge}
\end{figure}

Figure \ref{fig: civitation knowledge} illustrates how the local pressure changes in a one-dimensional flow. The valve would operate normally without cavitation when the minimum pressure ${p}_{min}$ is higher than the vapor pressure ${p}_{v}$. In the case where ${p}_{v}>{p}_{min}$ cavitation begins. However, it cannot be measured directly as the minimum pressure occurs downstream of the restriction. In practice, the cavitation coefficient ${X}_{FZ}$ is equal to the ratio of the external pressure difference to the internal pressure difference. It can be determined empirically by assuming that cavitation noise only begins when the minimum pressure ${p}_{min}$ is equal to the vapor pressure ${p}_{v}$. Therefore, the cavitation coefficient ${X}_{FZ}$ can be measured by the noise, which depends on the load of the valve. The equations for the cavitation coefficient ${X}_{FZ}$ and the operating pressure ratio ${X}_{F}$ are given below:
\begin{equation}
{X}_{FZ}=\frac{{p}_{u}-{p}_{d}}{{p}_{u}-{p}_{min}} \,\,\,\,\,\,\,\,\,\, {X}_{F}=\frac{{p}_{u}-{p}_{d}}{{p}_{u}-{p}_{v}}
\end{equation}
where ${p}_{u}$ is the upstream pressure, ${p}_{d}$ is the downstream pressure, ${p}_{min}$ is the minimum pressure in the valve and ${p}_{v}$ is the vapor pressure. When all coefficients are known over the full opening range of the valve, the following statements can be made.

\begin{itemize}
\item ${X}_{F}<{X}_{FZ}$: The valve operates without cavitation and the flow is only turbulent or laminar.
\item ${X}_{F}\geq {X}_{FZ}$: For ${X}_{F}={X}_{FZ}$, the valve operates with incipient cavitation. As the difference between ${X}_{FZ}$ and ${X}_{F}$ increases, the cavitation zone grows as the pressure drops due to increasing flow velocities.
\item ${X}_{F}>1$: Here the bubbles do not implode in the valve but rather continue to flow into the pipe because the downstream pressure ${p}_{d}$ is lower than the vapor pressure ${p}_{v}$. This phenomenon is called flashing.
\end{itemize}
The cavitation coefficient ${X}_{FZ}$ is only applied to the fluid, where it is measured empirically. Its valve varies for different liquid mediums, due to changes in viscosity, content of dissolved gas and so on.
\begin{figure}[htbp]
    \centering
    \includegraphics[width=0.45\textwidth,height=30mm]{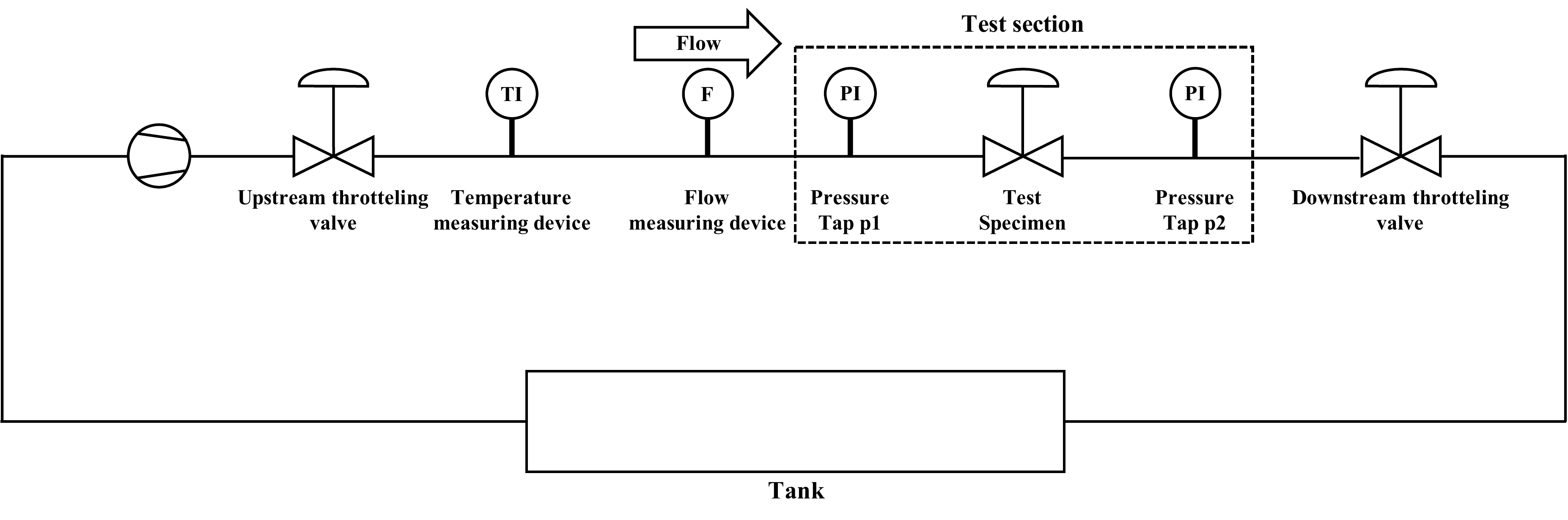}
    \caption{Schematic view of the test rack at SAMSON AG.}
    \label{fig: system}
\end{figure}
\begin{figure}[htbp]
    \centering
     \includegraphics[width=0.35\textwidth,height=38mm]{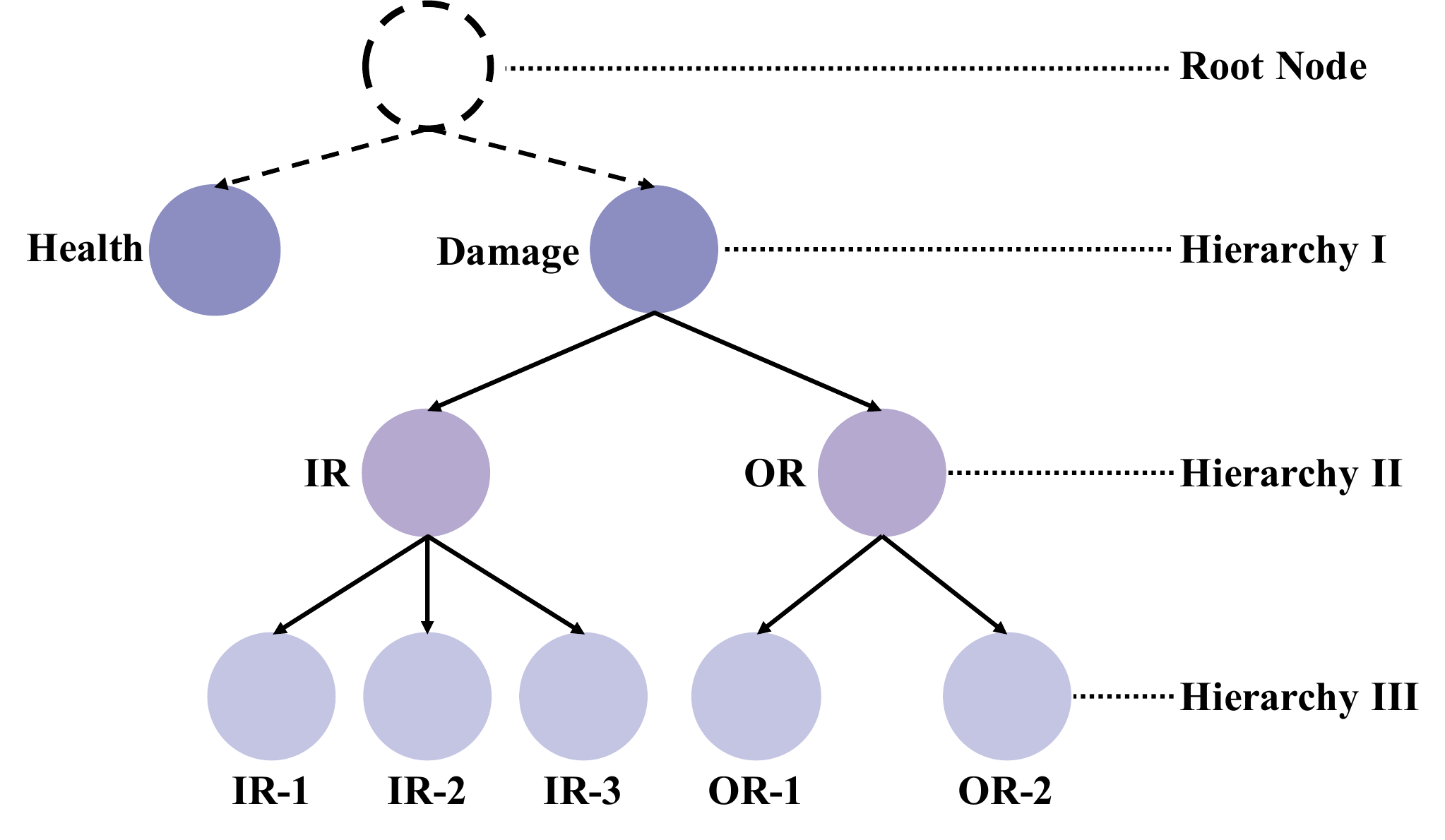}
    \caption{A hierarchical bearing tree from the PUB.}
    \label{fig: Appendix PUBD Hierarchical Tree}
\end{figure}
\begin{figure*}[htbp]
\subfigure[Swin-B (Cavitation-Short)]{
\begin{minipage}[t]{0.23\linewidth}
\centering
\includegraphics[width=\textwidth,height=30mm]{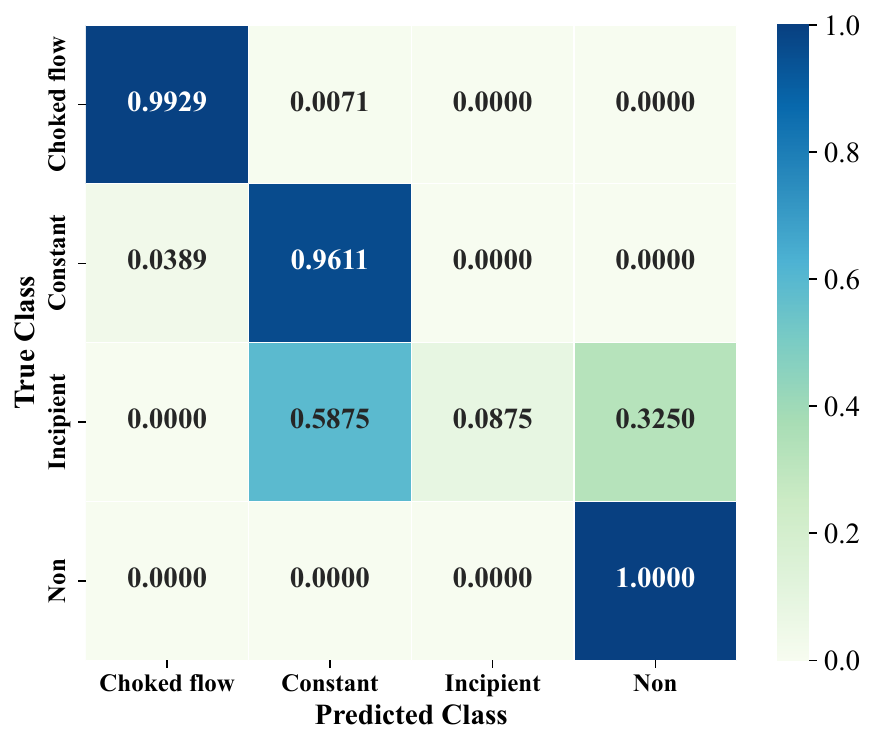}
%\caption{fig1}
\end{minipage}%
}%
\subfigure[Swin-B (Cavitation-Long)]{
\begin{minipage}[t]{0.23\linewidth}
\centering
\includegraphics[width=\textwidth,height=30mm]{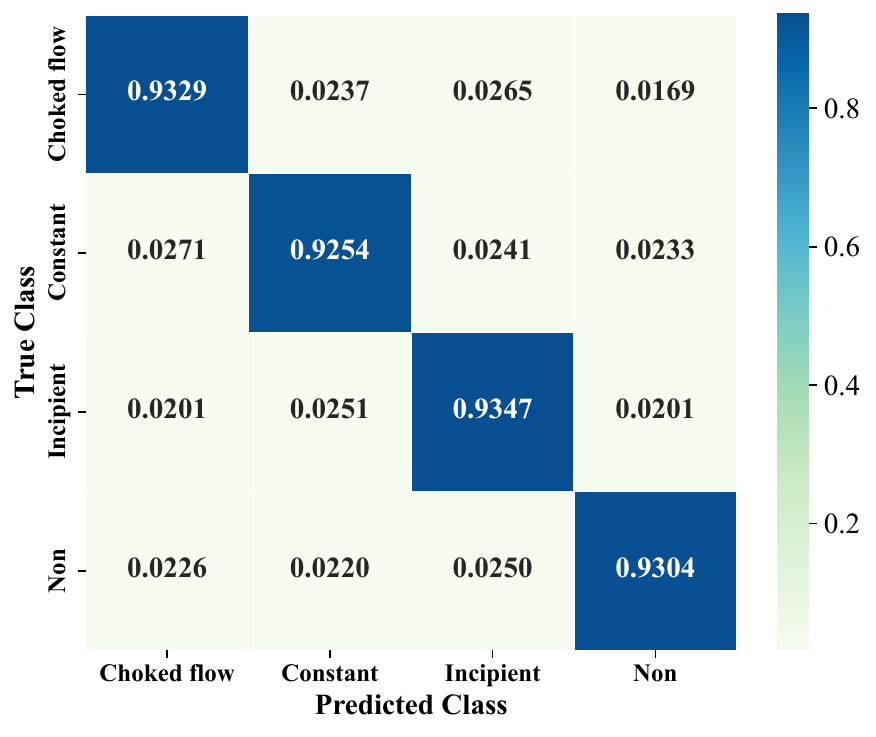}
%\caption{fig1}
\end{minipage}%
}%
\subfigure[Swin-B (Cavitation-Noise)]{
\begin{minipage}[t]{0.23\linewidth}
\centering
\includegraphics[width=\textwidth,height=30mm]{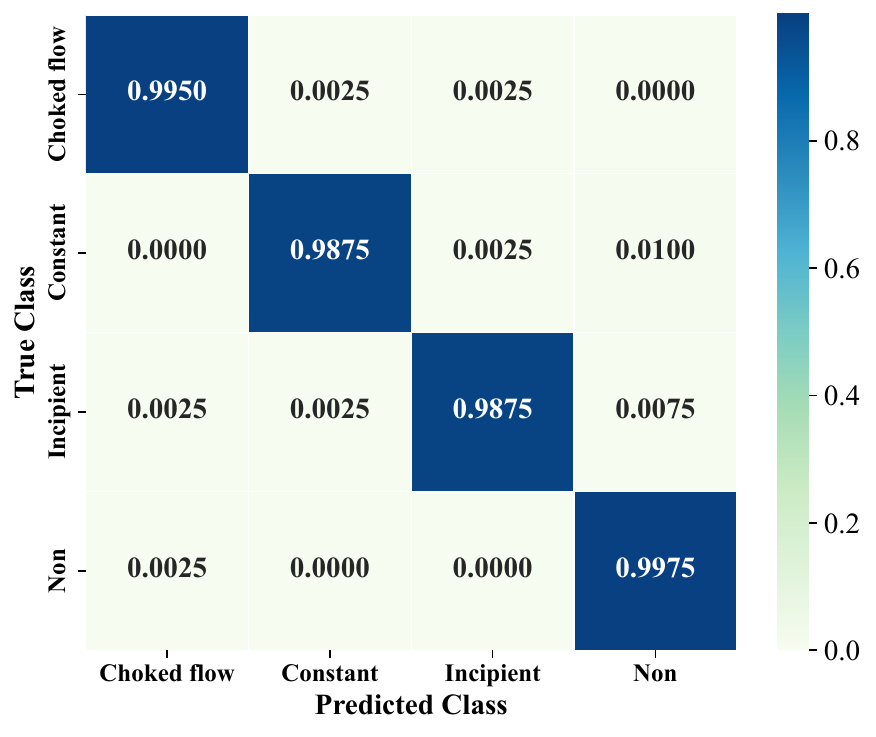}
%\caption{fig1}
\end{minipage}%
}%
\subfigure[TS-TCC (PUB)]{
\begin{minipage}[t]{0.23\linewidth}
\centering
\includegraphics[width=\textwidth,height=30mm]{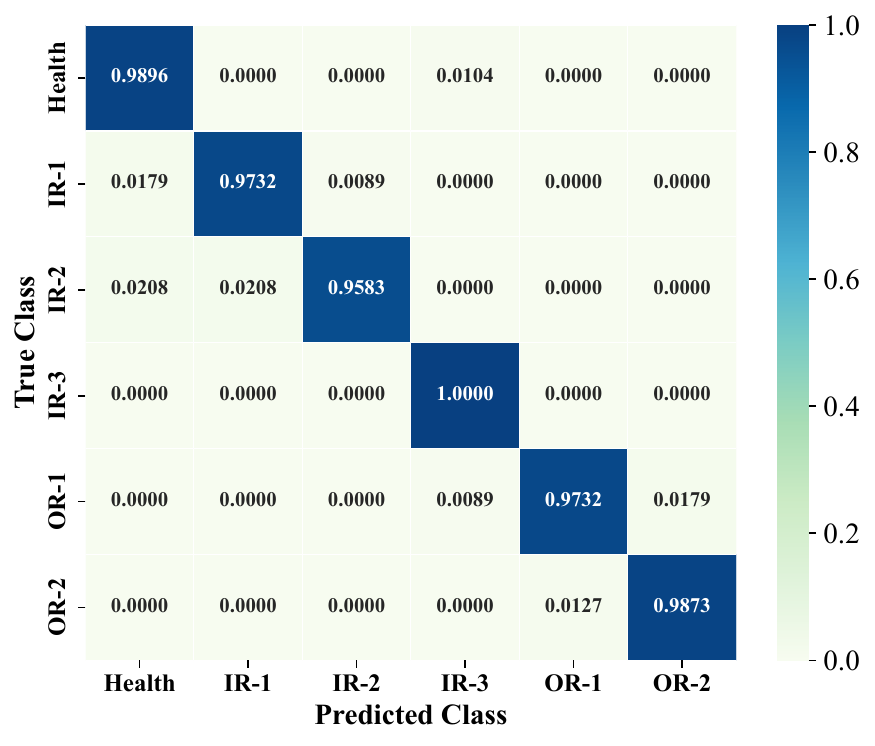}
%\caption{fig1}
\end{minipage}%
}%

\subfigure[HKG+Swin-B (Cavitation-Short)]{
\begin{minipage}[t]{0.23\linewidth}
\centering
\includegraphics[width=\textwidth,height=30mm]{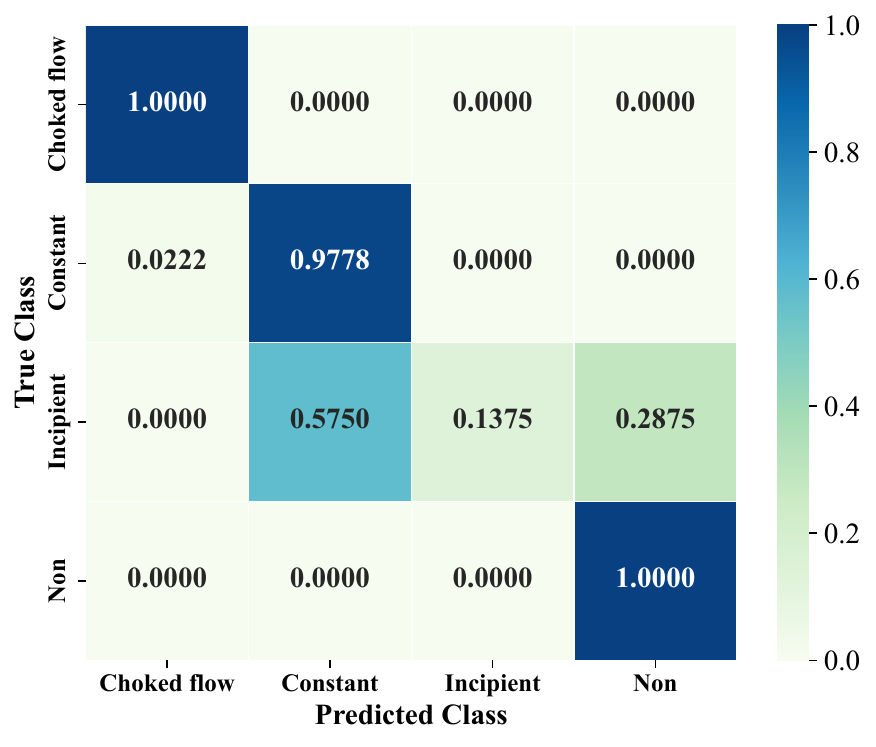}
%\caption{fig1}
\end{minipage}%
}%
\subfigure[HKG+Swin-B (Cavitation-Long)]{
\begin{minipage}[t]{0.23\linewidth}
\centering
\includegraphics[width=\textwidth,height=30mm]{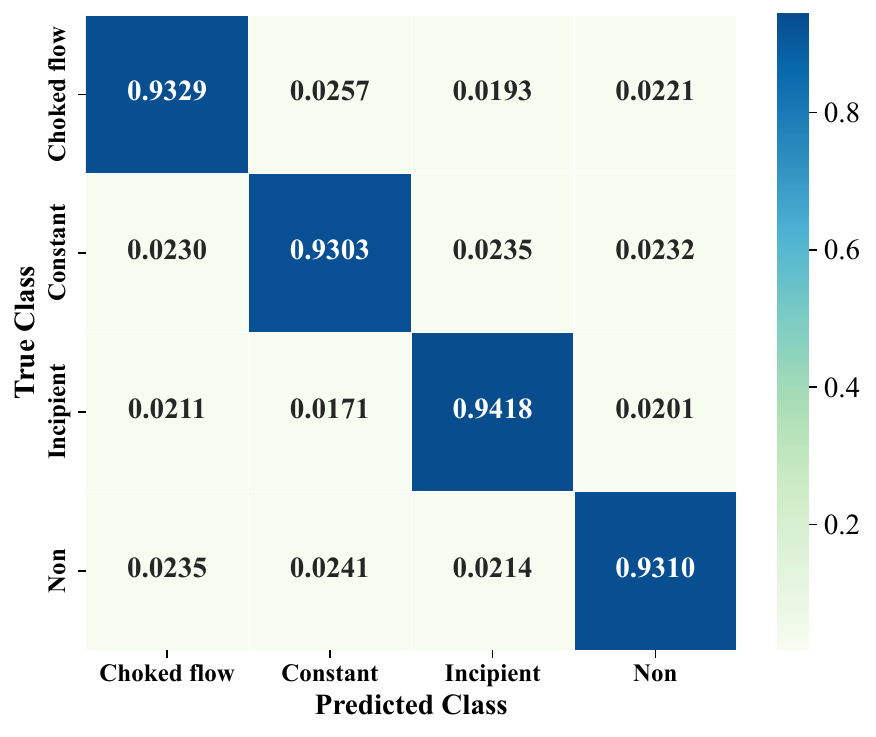}
%\caption{fig1}
\end{minipage}%
}%
\subfigure[HKG+Swin-B (Cavitation-Noise)]{
\begin{minipage}[t]{0.23\linewidth}
\centering
\includegraphics[width=\textwidth,height=30mm]{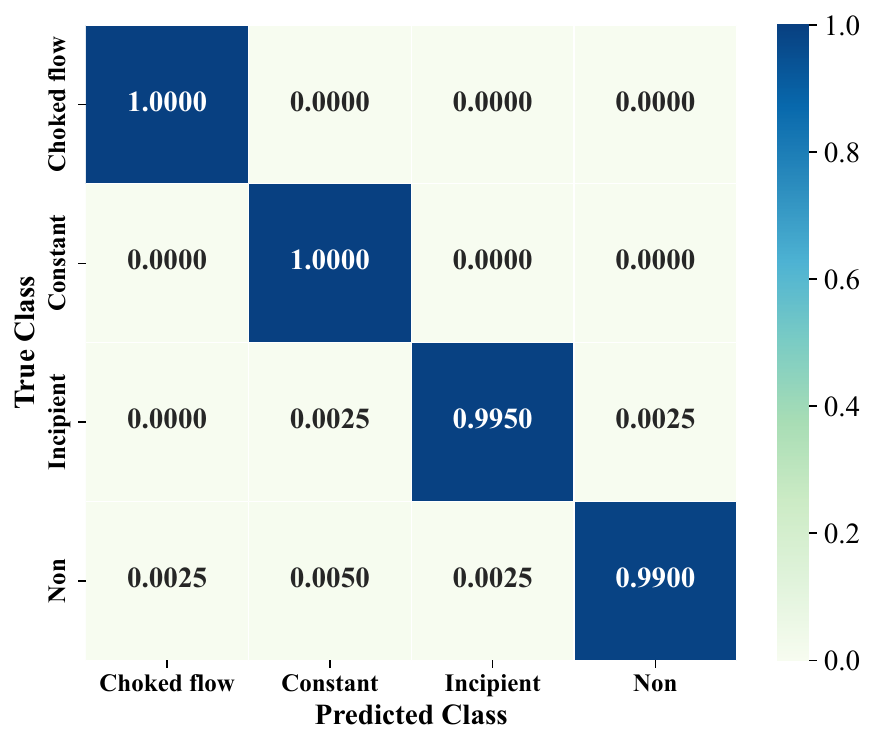}
%\caption{fig1}
\end{minipage}%
}%
\subfigure[HKG+ViT-S (PUB)]{
\begin{minipage}[t]{0.23\linewidth}
\centering
\includegraphics[width=\textwidth,height=30mm]{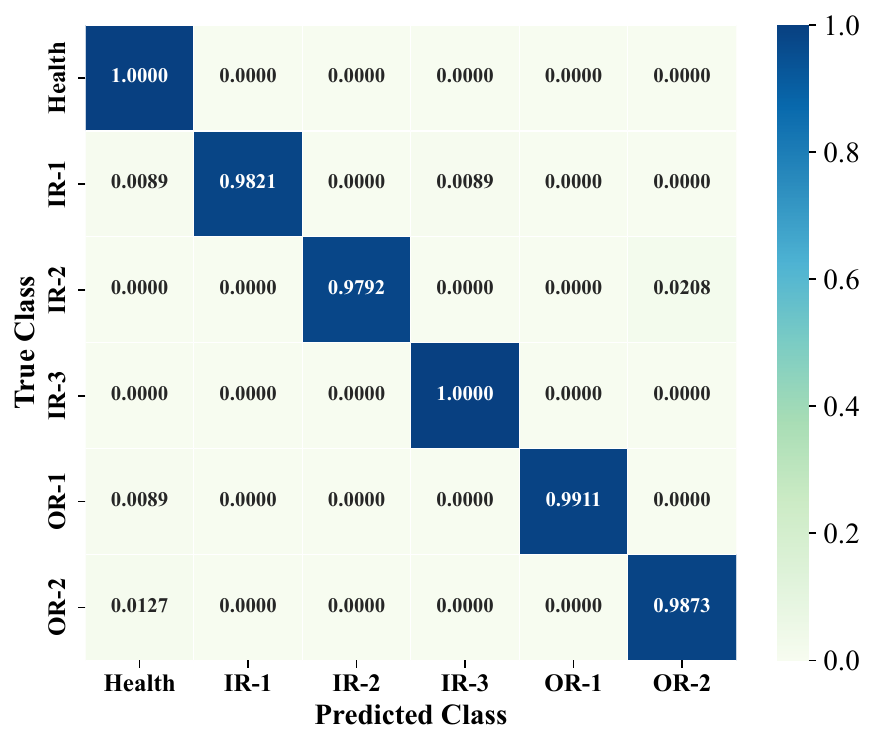}
%\caption{fig1}
\end{minipage}%
}%
\centering
\caption{The confusion matrix of HKG with different backbone networks on different datasets. (a)-(c) and (e)-(g) denote the confusion matrix on Cavitation-Short, Cavitation-Long and Cavitation-Noise, respectively. (d) and (h) on the PUB dataset.} 
\label{fig: Appendix Confusion Matrix}
\end{figure*}

\section{Experiments}
\label{sec: Appendix Experiments}
\subsection{Evaluation Metrics}
\label{sec: Appendix Evaluation Metrics}
As discussed in the evaluation metrics, we use dynamic thresholds to calculate the performance of fault intensity diagnosis for the proposed HKG. Therefore, given a specific threshold, we can calculate the TP (True Positives), FP (False Positives), TN (True Negatives), and FN (False Negatives). Furthermore, we have:
\begin{equation}
\begin{aligned}
\mathrm{Accuracy} &= \frac{\mathrm{TP} + \mathrm{TN}}{\mathrm{TP} + \mathrm{TN} + \mathrm{FP} + \mathrm{FN}}
\,\,\,\,\,\,\mathrm{Precision} &= \frac{\mathrm{TP}}{\mathrm{TP} + \mathrm{FP}} \\
\\
\mathrm{F1\text{-}score} &= \frac{2 \times \mathrm{Precision} \times \mathrm{Recall}}{\mathrm{Precision} + \mathrm{Recall}}
\,\,\,\,\,\,\mathrm{Recall} &= \frac{\mathrm{TP}}{\mathrm{TP} + \mathrm{FN}}\\
\end{aligned}
\end{equation}    
where under all possible thresholds, we can obtain a precision-recall curve (with precision as the $y$-axis and recall as the $x$-axis) and $AP = \sum\nolimits_n {(({R_n} - {P_{n - 1}})/{P_n}} )$, where ${P}_{n}$ and ${R}_{n}$ are precision and recall at the $n$-th threshold. 

\subsection{Datasets}
\label{sec: Appendix Datasets}
\noindent\textbf{Cavitation Datasets.} The cavitation datasets are provided by SAMSON AG in Frankfurt. The schematic of the experimental setup to collect data is shown in Figure \ref{fig: system}. And five flow statuses are induced in acoustic signals by varying the differential pressure at various constant upstream pressures of the control valve different operation conditions: choked flow cavitation, constant cavitation, incipient cavitation, turbulent flow and no flow (see Table \ref{tab: CavitationDatasets-FlowStatus} and Table \ref{tab: CavitationDatasets-operation}). For detailed dataset statistics shown in Table \ref{tab: training and test sets}. 
\begin{table}[htbp]
\centering
\caption{Cavitation datasets content details.}
\label{tab: CavitationDatasets-FlowStatus}
% \scriptsize
\footnotesize
\setlength{\tabcolsep}{0.5mm}{
\begin{tabular}{lccccc}
\toprule 
\multirow{2}{*}{Dataset} & \multicolumn{3}{c}{Cavitation} & \multicolumn{2}{c}{Non-cavitation}  \\ 
\cmidrule{2-6} 
& \multicolumn{1}{l}{choked flow} & \multicolumn{1}{l}{constant} & \multicolumn{1}{l}{incipient} & \multicolumn{1}{l}{turbulent} & \multicolumn{1}{l}{no flow} \\ 
\midrule
Cavitation-Short           & 72    & 93   & 40   & 118   & 33  \\ 
Cavitation-Long            & 148   & 396  & 64   & 183   & 15  \\ 
Cavitation-Noise           & 40    & 40   & 40   & 40    & 0  \\ 
\bottomrule
\end{tabular}}
\end{table}

\begin{table}[htbp]
\centering
\caption{Details of cavitation datasets valve operation.}
\label{tab: CavitationDatasets-operation}
% \scriptsize
\footnotesize
\setlength{\tabcolsep}{0.1mm}{
\begin{tabular}{lccc}
\toprule 
\multicolumn{1}{c}{\multirow{2}{*}{Dataset}} & \multicolumn{3}{c}{Operation parameters} \\ 
\cmidrule{2-4} 
\multicolumn{1}{c}{}    & \multicolumn{1}{c}{\begin{tabular}[c]{@{}c@{}}Valve stroke\\ (\SI{}{\mm})\end{tabular}} 
                        & \multicolumn{1}{c}{\begin{tabular}[c]{@{}c@{}}Upstream pressure\\ (\SI{}{\bar})\end{tabular}} 
                        & \multicolumn{1}{c}{\begin{tabular}[c]{@{}c@{}}Temperature\\ (\SI{}{\degreeCelsius})\end{tabular}} \\ 
\midrule
Cavitation-Short        & {[}15,13.5,11.25,7.5,3.75,1.5,0.75{]}      & {[}10,9,6,4{]}        & 25-50       \\
Cavitation-Long         & {[}60,55,45,30,25,15,6{]}                  & {[}10,6,4{]}          & 23-52        \\
Cavitation-Noise        & 15                                         & 10                    & 32-39         \\ 
\bottomrule
\end{tabular}}
\end{table}
\begin{table}[htbp]
\caption{Details of the training and test sets. $(\cdot )$ denotes the number after the sliding window (window size is 466944).}
\label{tab: training and test sets}
\scriptsize
\centering
\setlength{\tabcolsep}{0.1mm}{
\begin{tabular}{lcccccccc}
\toprule
                 \multicolumn{1}{c}{\multirow{3}{*}{Dataset}} & \multicolumn{4}{c}{Training set}      & \multicolumn{4}{c}{Testing set} \\ 
\cmidrule{2-9} 
                 & \multicolumn{3}{c}{Cavitation}       
                 & \multicolumn{1}{c|}{\multirow{2}{*}{Non}}
                 & \multicolumn{3}{c}{Cavitation}     
                 & \multirow{2}{*}{Non} \\ 
\cmidrule{2-4}
\cmidrule{6-8}
                 & Choked flow & Constant   & Incipient & \multicolumn{1}{c|}{}     & Choked flow & Constant & Incipient &    \\ 
\midrule
Cavitation-Short & 58($\times$10)    & 75($\times$10)   & 32($\times$10)  & \multicolumn{1}{c|}{121($\times$10)}  & 14($\times$10)    & 18($\times$10) & 8($\times$10)    & 30($\times$10)    \\
Cavitation-Long  & 118($\times$83)   & 317($\times$83) & 52($\times$83)  & \multicolumn{1}{c|}{158($\times$83)}  & 30($\times$83)    & 79($\times$83) & 12($\times$83)  & 40($\times$83)    \\
Cavitation-Noise & 32($\times$83)    & 32($\times$83)   & 32($\times$83)  & \multicolumn{1}{c|}{32($\times$83)}    & 8($\times$83)      & 8($\times$83)   & 8($\times$83)    & 8($\times$83)      \\ 
\bottomrule
\end{tabular}
}
\end{table}

\noindent\textbf{PUB Dataset.} This dataset is used to validate the extensibility of our approach. The bearing damage levels are listed in Table \ref{tab: damage levels PUBD}. The bearing file codes and fault types utilized in our experiments are shown in Table \ref{tab: fault type and file code}. The PUB is organized into three hierarchies: bearing diagnosis (Hierarchy I), bearing damage type diagnosis (Hierarchy II), and bearing IR/OR intensity diagnosis (Hierarchy III-IR/III-OR), as shown in Figure \ref{fig: Appendix PUBD Hierarchical Tree}. 
\begin{table}[htbp]
\centering
\caption{Bearing fault damage levels on the PUB.}
\label{tab: damage levels PUBD}
% \scriptsize
\footnotesize
\begin{tabular}{ccc}
\toprule
Damage level & Percentage values  & Bearing limitations   \\ 
\midrule
1            & 0-2$\%$              & $\le$\SI{2}{mm}      \\
2            & 2-5$\%$              & $>$\SI{2}{mm}    \\
3            & 5-15$\%$             & $>$\SI{4.5}{mm}\\
\bottomrule
\end{tabular}
\end{table}

\begin{table}[htbp]
\caption{Bearing fault types and file codes on the PUB.}
\label{tab: fault type and file code}
% \scriptsize
\footnotesize
\centering
\begin{tabular}{ccccccc}
\toprule
\multirow{2}{*}{Fault type} & \multirow{2}{*}{Healthy} & \multicolumn{2}{c}{OR damage} & \multicolumn{3}{c}{IR damage} \\
\cmidrule{3-7} 
\multicolumn{1}{c}{}                            &                          & OR-1          & OR-2          & IR-1     & IR-2     & IR-3    \\
\midrule
\multicolumn{1}{l}{\multirow{7}{*}{File code}} & K001                     & KA01          & KA03          & KI01     & KI07     & KI16    \\
\multicolumn{1}{l}{}                           & K002                     & KA05          & KA06          & KI03     & KI08     & -       \\
\multicolumn{1}{l}{}                           & K003                     & KA04          & KA08          & KI04     & KI18     & -       \\
\multicolumn{1}{l}{}                           & K004                     & KA07          & KA09          & KI05     & -        & -       \\
\multicolumn{1}{l}{}                           & K005                     & KA15          & KA16          & KI14     & -        & -       \\
\multicolumn{1}{l}{}                           & K006                     & KA22          & -             & KI17     & -        & -       \\
\multicolumn{1}{l}{}                           & -                        & KA30          & -             & KI21     & -        & -       \\
\bottomrule
\end{tabular}
\end{table}

\subsection{Results}
\label{sec Appendix Results}
The confusion matrix of HKG with different backbone networks on four real-world datasets, see Figure \ref{fig: Appendix Confusion Matrix}.

\end{document}